\documentclass[10pt,journal,compsoc]{IEEEtran}

\ifCLASSOPTIONcompsoc
  \usepackage[nocompress]{cite}
\else
  \usepackage{cite}
\fi

\ifCLASSINFOpdf
   \usepackage[pdftex]{graphicx}
   \graphicspath{{../pdf/}{../jpeg/}}
   \DeclareGraphicsExtensions{.pdf,.jpeg,.png}
\else
   \usepackage[dvips]{graphicx}
   \graphicspath{{../eps/}}
   \DeclareGraphicsExtensions{.eps}
\fi
\usepackage[numbers]{natbib}
\usepackage{color}
\usepackage{amsthm}
\usepackage{bbm}
\usepackage{booktabs}
\usepackage{algorithm}
\usepackage{algorithmic}
\usepackage{graphicx}
\usepackage{color}
\usepackage{amsmath}
\usepackage{amssymb}
\usepackage{amsfonts}
\usepackage{multirow}
\usepackage{epsfig}
\newcommand{\mName}{TriDA }
\usepackage{graphicx}
\usepackage{courier}  
\usepackage[hyphens]{url}  
\usepackage[table,xcdraw]{xcolor}
\usepackage[colorlinks,linkcolor=blue]{hyperref}
\usepackage{xspace}
\usepackage{textcomp}
\usepackage{xspace}
\newcommand{\etal}{\textit{et al.}\@\xspace}
\newcommand{\ie}{\textit{i.e.}\@\xspace}
\newcommand{\eg}{\textit{e.g.}\@\xspace}
\usepackage{pifont}

 \usepackage[capitalize]{cleveref}
    \crefname{section}{Sec.}{Secs.}
    \Crefname{section}{Section}{Sections}
    \Crefname{table}{Table}{Tables}
    \crefname{table}{Tab.}{Tabs.}

\theoremstyle{plain}

\newcommand{\revise}[1]{\textcolor{black}{#1}}
\newtheorem{Definition}{Definition}
\newtheorem{Proposition}{Proposition}
\newtheorem*{Finding}{Finding}
\newtheorem{Factor}{Factor}

\theoremstyle{remark}
\newtheorem*{Discussion}{Discussion}
\begin{document}

\title{Incorporating Pre-training Data Matters in Unsupervised Domain Adaptation}

\author{Yinsong Xu,
        Aidong Men,
        Yang Liu,
        Xiahai Zhuang~\textsuperscript{\ding{41}},
        Qingchao Chen~\textsuperscript{\ding{41}}
\IEEEcompsocitemizethanks{
\IEEEcompsocthanksitem This work was supported by grants from the National Natural Science Foundation of China (62201014). Clinical Medicine Plus X - Young Scholars Project of Peking University(PKU2024LCXQ028), the Fundamental Research Funds for the Central Universities.

\IEEEcompsocthanksitem Yinsong Xu is with the National Institute of Health Data Science, also the Institute of Medical Technology, Peking University, Beijing, 100191, China. E-mail: xuyinsong@bupt.edu.cn
\IEEEcompsocthanksitem Aidong Men is with the School of Artificial Intelligence, Beijing University of Posts and Telecommunications, Beijing 100876, China. E-mail: menad@bupt.edu.cn.
\IEEEcompsocthanksitem Yang Liu is with Wangxuan Institute of Computer Technology, Peking University, Beijing, 100080, China, E-mail: yangliu@pku.edu.cn.
\IEEEcompsocthanksitem Xiahai Zhuang is with the School of Data Science, Fudan University, Shanghai, 200433, China. (e-mail: zxh@fudan.edu.cn).
\IEEEcompsocthanksitem Qingchao Chen is with the National Institute of Health Data Science and also the Institute of Medical Technology, Peking University, Beijing, 100191, China. He is also with the State Key Laboratory of General Artificial Intelligence, Peking University. E-mail: qingchao.chen@pku.edu.cn.
\IEEEcompsocthanksitem \ding{41} Corresponding author: Qingchao Chen and Xiahai Zhuang.}
\thanks{Manuscript received April 19, 2005; revised August 26, 2015.}}

\markboth{Journal of \LaTeX\ Class Files,~Vol.~14, No.~8, August~2021}%
{Shell \MakeLowercase{\textit{et al.}}: A Sample Article Using IEEEtran.cls for IEEE Journals}

\IEEEpubid{0000--0000/00\$00.00~\copyright~2021 IEEE}

\IEEEtitleabstractindextext{

\begin{abstract}
In deep learning, initializing models with pre-trained weights has become the \textit{de facto} practice for various downstream tasks. Many unsupervised domain adaptation (UDA) methods typically adopt a backbone pre-trained on ImageNet, and focus on reducing the source-target domain discrepancy. However, the impact of pre-training on adaptation received little attention. In this study, we delve into UDA from the novel perspective of pre-training. We first demonstrate the impact of pre-training by analyzing the dynamic distribution discrepancies between pre-training data domain and the source/ target domain during adaptation. Then, we reveal that the target error also stems from the pre-training in the following two factors: 1) empirically, target error arises from the gradually degenerative pre-trained knowledge during adaptation; 2) theoretically, the error bound depends on difference between the gradient of loss function, \ie, on the target domain and pre-training data domain. To address these two issues, we redefine UDA as a three-domain problem, \ie, source domain, target domain, and pre-training data domain; then we propose a novel framework, named \textbf{TriDA}. We maintain the pre-trained knowledge and improve the error bound by incorporating pre-training data into adaptation for both vanilla UDA and source-free UDA scenarios. For efficiency, we introduce a selection strategy for pre-training data, and offer a solution with synthesized images when pre-training data is unavailable during adaptation. Notably, \mName is effective even with a small amount of pre-training or synthesized images, and seamlessly complements the two scenario UDA methods, demonstrating state-of-the-art performance across multiple benchmarks. We hope our work provides new insights for better understanding and application of domain adaptation. Code is available at \href{https://github.com/SPIresearch/TriDA.git}{https://github.com/SPIresearch/TriDA.git}.

\end{abstract}

\begin{IEEEkeywords}
Unsupervised domain adaptation, pre-training, distribution discrepancy.
\end{IEEEkeywords}}

\maketitle

\section{Introduction}

\IEEEPARstart{D}{eep} neural networks often suffer from the \textit{domain shift} when evaluated on out-of-distribution (OOD) test data \cite{gulrajanisearch,torralba2011unbiased}. Unsupervised domain adaptation (UDA) emerges as a promising way to address it, which transfers knowledge from the label-rich source domain to the unlabeled target domain \cite{xiao2021dynamic, Lin2022ProCA, Chen_2018_CVPR}. 

Ben-David \etal \cite{ben2010theory} provide an upper bound on the \emph{target error} as the sum of the source error, source-target domain discrepancy, and a constant \cite{ben2010theory, zhang2019bridging}. Consequently, existing methods formulate UDA as a problem involving two domains: the source and target domains, and follow the well-established two-domain paradigm, reducing target error by minimizing the source-target domain discrepancy \cite{ganin2015unsupervised, long2018conditional,zhang2019bridging}. 

However, the two-domain paradigm tends to overlook a critical aspect: pre-training, despite its popularity and general efficacy. In deep learning, initializing models with pre-trained weights (\eg, trained on ImageNet \cite{deng2009imagenet}) has been the \textit{de facto} practice across various downstream tasks \cite{zhou2022rethinking, li2022deep, simonyan2014two, li2022exploring}. It accelerates convergence \cite{he2019rethinking}, and offers performance benefits based on the alignment between the pre-training dataset and downstream tasks \cite{raghu2019transfusion}. However, under the domain shift between the source and target domain in the UDA, \emph{the exact role of the pre-training has not been fully investigated}, either empirically or theoretically. In this paper, as shown in \cref{fig:roadmap}, we explore it by answering the following three questions:

\begin{figure}[t]
    \centering
    \includegraphics[width=0.8\linewidth]{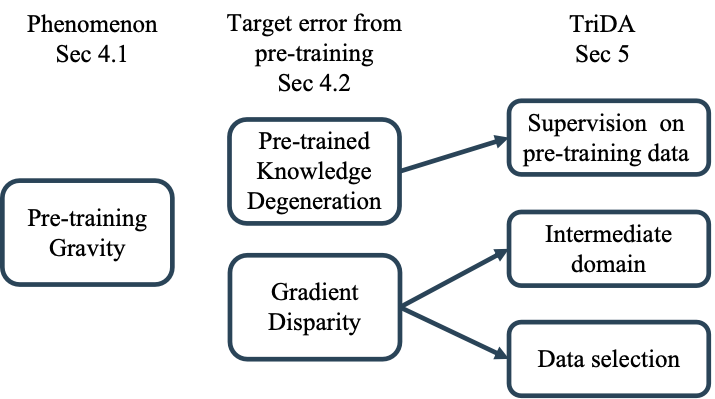}
    \caption{Roadmap: Our TriDA improve the target error stemming from
pre-training.}
    \label{fig:roadmap}
\end{figure}

\begin{figure*}
    \centering
\includegraphics[width=\linewidth]{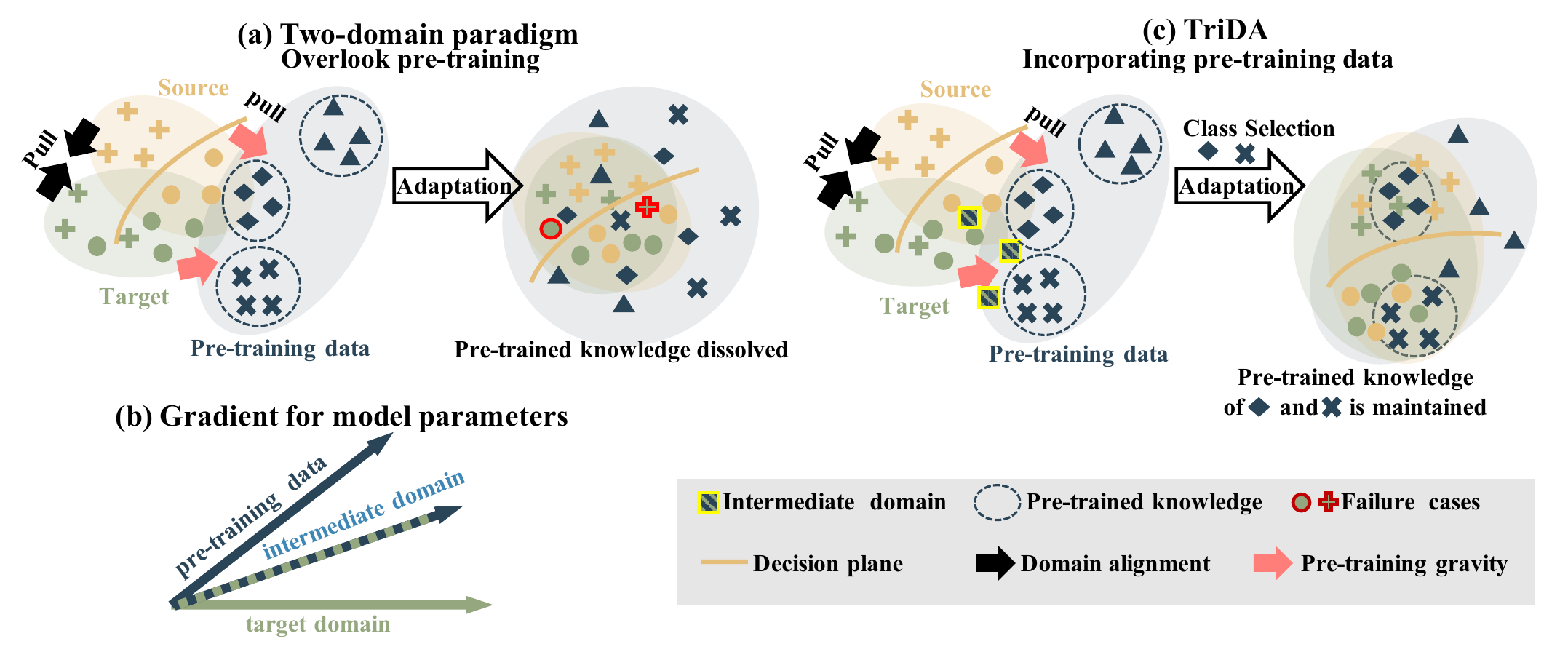}
    \caption{(Best viewed in color) The impact of pre-training in domain adaptation. (a) We observe an interesting phenomenon termed \textbf{\textit{Pre-training Gravity}}: the {{source}} and {target} features are \textit{pulled} towards the features of pre-training data. \textbf{\textit{Pre-trained Knowledge Degeneration}} (\ie dotted circles disappeared) increases the target error during the alignment between the target domain and pre-training data due to pre-training gravity. (b) We term the difference in gradient of model parameters when replacing the domain during optimization as \textbf{\textit{Gradient Disparity}}. Our theoretical analysis demonstrates that target error has a dependency on the gradient disparity between the target domain and the pre-training data. (c) TriDA addresses the two issues by 1) incorporating selected pre-training data to prevent pre-trained knowledge degeneration, and 2) introducing the intermediate domain to align the gradient for model parameters with pre-training data.}
    \label{teaser}
\end{figure*}

\textbf{Does pre-training impact adaptation?} 
\textit{Yes. Not only the source knowledge, but pre-trained knowledge is also transferred to the target domain during adaptation.} The impact is most directly reflected in performance, which varies across different pre-training methods, network architectures, and datasets \cite{kim2022broad}. Furthermore, we delve into the dynamic evolution of distribution discrepancies among pre-training data, source, and target domains. Intuitively, during adaptation, the source knowledge is transferred to the target domain, resulting in the distribution alignment between the source features and target features. More importantly, as shown in \cref{teaser}, we observe an interesting phenomenon termed \textbf{\textit{Pre-training Gravity}}: the source and target features are \textit{pulled} towards the features of pre-training data. Specifically, the distribution discrepancies between pre-training data and source/target gradually diminish during adaptation without any procedure related to pre-training in the adaptation, highlighting the impact of pre-training.


\textbf{How does pre-training impact target error?} \textit{We summarize its impact into two critical factors: Pre-trained Knowledge Degeneration and Gradient Disparity.} Firstly, we quantify the pre-trained knowledge using silhouette scores during the adaptation and observe a phenomenon termed \textbf{\textit{Pre-trained Knowledge Degeneration}}: during adaptation, pre-trained knowledge undergoes degeneration when dealing with the inevitable noisy target pseudo-labels caused by the bias of pre-trained representations \cite{chen2022debiased}. Unlike catastrophic forgetting \cite{kirkpatrick2017overcoming}, where models tend to forget previously learned knowledge when exposed to new data or tasks, pre-trained knowledge degeneration does \textit{not} occur when using ground-truth as supervision. This reveals that the target error arises due to the degenerative pre-trained knowledge being transferred to the target domain, \ie, the distribution alignment between the target features and the degraded pre-trained features, as shown in \cref{teaser}(a). Secondly, we conduct a theoretical analysis through the lens of model optimization with stochastic gradient descent. We term the difference in gradient of model parameters as \textbf{\textit{Gradient Disparity}} when replacing the domain during optimization (details refer to \cref{how}), as shown in \cref{teaser}(b). Our theoretical analysis demonstrates that target error has a dependency on the gradient disparity between the target domain and the pre-training data. \textit{The two factors of target error stemming from pre-training are not considered by the two-domain paradigm, which solely aligns source and target domains.}

\textbf{How to improve the target error stemming from pre-training?} We extend the two-domain paradigm by redefining UDA as a \textit{three-domain} problem: pre-training data, source, and target domains. Then, we present a novel framework \textbf{TriDA}, as shown in \cref{teaser}(c). Firstly, to prevent the degeneration of pre-trained knowledge, we reuse the pre-training data and utilize its supervision as an intervention of the semantic knowledge during the learning on the target domain. Secondly, we demonstrate that the target error bound is influenced by two key factors: the objective function applied to the pre-training data and the reuse of pre-training data itself. To address the first factor, \ie the objective function, we impose constraints on the intermediate domain, which is formed by interpolating between the target and pre-training data domains; to address the second factor, the reuse of pre-training data, we present a class-based data selection strategy, which also helps reduce the number of incorporated images, thereby improving efficiency. As the pre-training dataset may be \textit{unavailable} due to data privacy, security, and proprietary restrictions, we also present a solution with synthesized images for such scenario. Our method proves effective even with a small amount of pre-training or synthesized images.

This paper investigates domain adaptation from a novel perspective, focusing on the impact of pre-training and addressing the target error arising from it, rather than solely improving the alignment between source and target domains. Consequently, our method can seamlessly enhance existing vanilla UDA and Source-Free Unsupervised Domain Adaptation (SFUDA) methods. Our contributions are summarized as follows:
\begin{itemize}

    \item  We highlight the phenomenon Pre-training Gravity. Through empirical and theoretical analyses, we reveal the mechanisms of target error stemming from pre-training, identifying two key issues: Pre-trained Knowledge Degeneration and Gradient Disparity. To the best of our knowledge, these problems have not been addressed by conventional two-domain UDA methods.



    \item We redefine UDA as a \textit{three-domain} problem, and propose a novel framework, named TriDA. It incorporates constraints on the intermediate domain and a data selection process to reduce the error bound.

    \item Our proposed method can be seamlessly integrated with existing UDA and SFUDA techniques, achieving state-of-the-art results across multiple benchmarks, demonstrating its scalability. It also offers a solution when pre-training data is unavailable, highlighting its viability. 

    \item Our work provides valuable insights into the impact of pre-training on UDA, offering new strategies for effectively adapting pre-trained models to downstream tasks.
    
\end{itemize}

\section{Related Work}
\subsection{Pre-training and Fine-tuning.} 
The "pre-training $\rightarrow$ fine-tuning" paradigm, widely adopted in transfer learning, has achieved state-of-the-art results in many computer vision applications \cite{li2022video,nguyen2022boxer} thanks to large-scale datasets \cite{deng2009imagenet, yang2022s} and public pre-trained models offered by PyTorch, TensorFlow, and third-party libraries. In spite of the immense popularity, some recent works question the paradigm. He et al. \cite{he2019rethinking} observe that training from random initialization can yield comparable results to pre-trained models across a range of baseline tasks. Empirical results suggest that it offers diminishing advantages for localization tasks \cite{he2019rethinking} and fine-grained classification \cite{kornblith2019better}. Meanwhile, a group of works focuses on how to better utilize the pre-trained knowledge from the source dataset \cite{xuhong2018explicit, wortsman2022robust, azizi2021big, you2020co}. Liu \etal~\cite{liu2022improved} reuse pre-training data in fine-tune to prove the downstream performances. Kim \etal~\cite{kim2022broad} conduct the analysis of pre-training for domain adaptation and generalization, and empirically show that adaptation performances can be improved with better pre-training. 

Zhang \etal~\cite{zhang2023rethinking} empirically discuss the role of pre-trained networks in SFUDA from the aspect of pseudo labels. They show that the choice of pre-trained model can improve robustness, while source-trained models produce unreliable pseudo labels on target samples. Consequently, they propose Co-learn, which distills information from pre-trained networks to improve target pseudo labels. In our study, we explore a different perspective by delving deeper into the impact of pre-training on target error. Through empirical and theoretical analyses, we identify two factors related to target error that stem from pre-training, revealing that adaptation is influenced not only by the source and target domains but also by the pre-training data.

\subsection{Vanilla Unsupervised Domain Adaptation.} In vanilla UDA, we have access to a labeled source and an unlabeled target dataset. Adversarial learning based methods \cite{goodfellow2014generative} typically employ a feature generator and a domain discriminator that play a minimax game to align the marginal \cite{zhou2023self, ganin2015unsupervised,li2020maximum} or conditional distribution \cite{long2018conditional}. BCDM \cite{Li21BCDM} encourages the bi-classifier to generate disagreement prediction to improve target features' categorical discriminability. ATDOC \cite{Liang_2021_CVPR} generates pseudo labels with class centroids. Another group of works propose distribution discrepancy measurements to promote domain confusion in the feature space, \eg  $\mathcal{H} \Delta \mathcal{H}$-divergence \cite{ben2010theory}, Maximum Classifier Discrepancy \cite{saito2018maximum} and Maximum Mean Discrepancy \cite{tolstikhin2016minimax}, Margin Disparity Discrepancy \cite{zhang2019bridging}, Maximum structural generation discrepancy
\cite{xia2022maximum}. In partial domain adaptation where the source label space subsumes the target one, IDSP~\cite{li2022partial} shows the target risk is bounded by both model smoothness and between-domain discrepancy, while giving up possibly riskier domain alignment.

\subsection{Source-free Unsupervised Domain Adaptation.} 
Different from Vanilla UDA, Source-free Unsupervised Domain Adaptation (SFUDA) \cite{liang2020we, ding2022source, pei2023uncertainty} assumes that the source data is unavailable during the adaptation. SFUDA scenarios involve unlabeled target data and a model pre-trained on source data. Source-free Unsupervised Domain Adaptation (SFUDA) \cite{liang2020we, liang2021source, ding2022source, pei2023uncertainty} scenarios involve unlabeled target data and a model pre-trained on source data. Pseudo-labeling \cite{liang2020we, lee2022confidence} serves as the mainstream framework, with methods like SHOT \cite{liang2020we} generating pseudo labels through clustering and incorporating information maximization and entropy minimization, and CoWA \cite{lee2022confidence} generating pseudo labels using Gaussian Mixture Modeling. When the source domain is unavailable, \cite{ding2022source} treats the classifier weights as centers of source features and then minimizes the discrepancy between the surrogate source and target distribution. Additionally, techniques such as local neighbor structure \cite{yang2022attracting, zhang2022divide, yang2021exploiting} also demonstrated performance benefits. Yang \etal~\cite{yang2023trust} explore the intrinsic neighborhood structure of the target data which have shifted in the feature space, and estimate the density around each data point and decrease the contribution of outliers on the clustering.

Previous UDA and SFUDA methods follow the two-domain paradigm that focuses on minimizing the source-target domain discrepancy, and pay little attention to pre-training on adaptation, particularly its impact on target error. Rather than improving the alignment between source and target domains, our work reveals the target error steaming from pre-training, which cannot be addressed by the two-domain paradigm. We address this by incorporating pre-training data and redefining UDA as a three-player problem. Finally, we introduce a novel framework, named TriDA, which can seamlessly complement existing methods.

\section{Background: The Two-Domain Paradigm}
\label{tdpara}

\textbf{Problem formulation. } The paradigm considers two domains: a labeled source domain $\mathcal{D}_s=\{(\textbf{x}_s^{(i)},\textbf{y}_s^{(i)})\}_{i=1}^{N_s}$, and an unlabeled target domain $\mathcal{D}_t=\{\textbf{x}_t^{(i)}\}_{i=1}^{N_t}$. The source domain and target domain are sampled from joint distributions $P_s(\textbf{X}_s, \textbf{Y}_s)$ and $P_t(\textbf{X}_t, \textbf{Y}_t)$, respectively, $P_s \neq P_t$. The aim is to learn a hypothesis $h$ parameterized by $\theta$ which predicts the labels in $\mathcal{D}_t$. 

\begin{figure}
    \centering
    \includegraphics[width=1\linewidth]{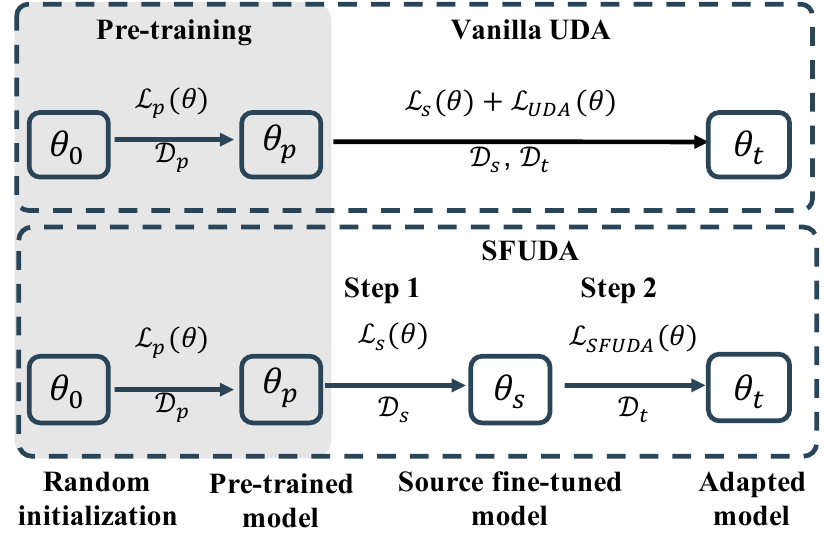}
    \caption{Two-domain paradigm methods focus on reducing target error by minimizing the source-target domain discrepancy and source error simultaneously in Vanilla UDA, or step-wise in SFUDA. They pay little attention to the pre-training (dashed box). $\theta_0$ is the random initialized model, and $\mathcal{L}_p$ is the objective in pre-training.  }
    \label{fig:enter-label}
\end{figure}

\begin{table}
\caption{Reference to the Mathematical Symbols}
\begin{center}
\begin{tabular}{ll}
    \toprule
    Symbols & Definition\\ \midrule
    $\mathcal{D}$ &dataset\\
    $P$ &data distributions\\
    $\textbf{x} / \textbf{y}$ &input data / label\\
    $\widetilde{\textbf{x}}$ & intermediate domain data\\
    $\hat{\textbf{y}}_t$  & target pseudo-label\\
    $\mathcal{L}_{s/ t /p}$ & objective function on the \\&source/ target/ pre-training domain\\
    $\theta_{0/ p/ s/ t }$ &random initialized / pre-trained /  \\ 
    & source fine-tuned / target-adapted \\ & resulting model\\
    $\mathcal{R}_t(\theta)$ &oracle cross-entropy on target domain\\
    $\widetilde{\mathcal{R}}_t(\theta)$ &target excess risk\\
    $\Delta(\cdot)$ &gradient disparity\\
    $\nabla$  &the gradient operator with respect to variable $\theta$\\
    $T$ & the number of optimization iterations\\
    $\mathcal{C}$ & the set of label classes\\
    $c_{(i)}$ & the $i$-th class in the classes set\\
    $\hat{\textbf{x}}_i$ & the synthesized data from label $c_{(i)}$\\

    \bottomrule
    \end{tabular}
\end{center}
\label{symbol}
\end{table}

\noindent \textbf{Optimization. }The target error of $h$ is given by $\epsilon_t(h) = \mathbb{E}_{(\textbf{x}_t,\textbf{y}_t)\sim P_t} \mathbbm{1} [h(\textbf{x}_t) \neq \textbf{y}_t]$, where $\mathbbm{1}[\cdot]$ is the indicator function. The seminal work \cite{ben2010theory} provides an upper bound on $\epsilon_t(h)$ defined as: 
\begin{align}
\label{bound} 
    \epsilon_t(h) \leq \epsilon_s(h) + \frac{1}{2}d_{\mathcal{H} \Delta \mathcal{H}}(P_s,P_t ) + \lambda,
\end{align}
where $\epsilon_s(h) = \mathbb{E}_{(\textbf{x}_s,\textbf{y}_s)\sim P_s} \mathbbm{1} [h(\textbf{x}_s) \neq \textbf{y}_s]$ is the source error, $d_{\mathcal{H} \Delta \mathcal{H}}(\cdot,\cdot)$ is the $\mathcal{H} \Delta \mathcal{H}$-divergence measuring the domain discrepancy, and $\lambda$ is the \emph{ideal combined error}, which is considered sufficiently small and related with the class discriminability \cite{xiao2021dynamic}. Building upon this foundation and subsequent theories \cite{zhang2019bridging, JMLR:v20:15-192}, as shown in \cref{fig:enter-label}, the two-domain paradigm solves UDA by reducing the first two items on the right hand side of \cref{bound}. Specifically, $\epsilon_s(h)$ is reduced by cross-entropy loss $\mathcal{L}_s$ on $\mathcal{D}_s$. 
$d_{\mathcal{H} \Delta \mathcal{H}}(\cdot,\cdot)$ is lowered by $\mathcal{L}_{UDA}$ on $\mathcal{D}_s$ with $\mathcal{D}_t$ simultaneously in Vanilla UDA~\cite{ganin2015unsupervised,long2018conditional, Li21BCDM}; or $\mathcal{L}_{SFUDA}$ on $\mathcal{D}_t$ step-wise in SFDUA \cite{liang2020we,lee2022confidence,yang2022attracting, yang2021exploiting,10420513}. \cref{symbol} provides the reference of the key symbols used in this paper.

\begin{Discussion}
    Existing two-domain paradigm works pay little attention to the pre-trained model $\theta_p$, which serves as the initialization of models and the key factor of the hypothesis space. Motivated by this, we delve into UDA from the perspective of pre-training data and analyze its impact in the subsequent sections.
\end{Discussion}

\begin{figure*}[t]
    \centering
    \includegraphics[width=1\linewidth]{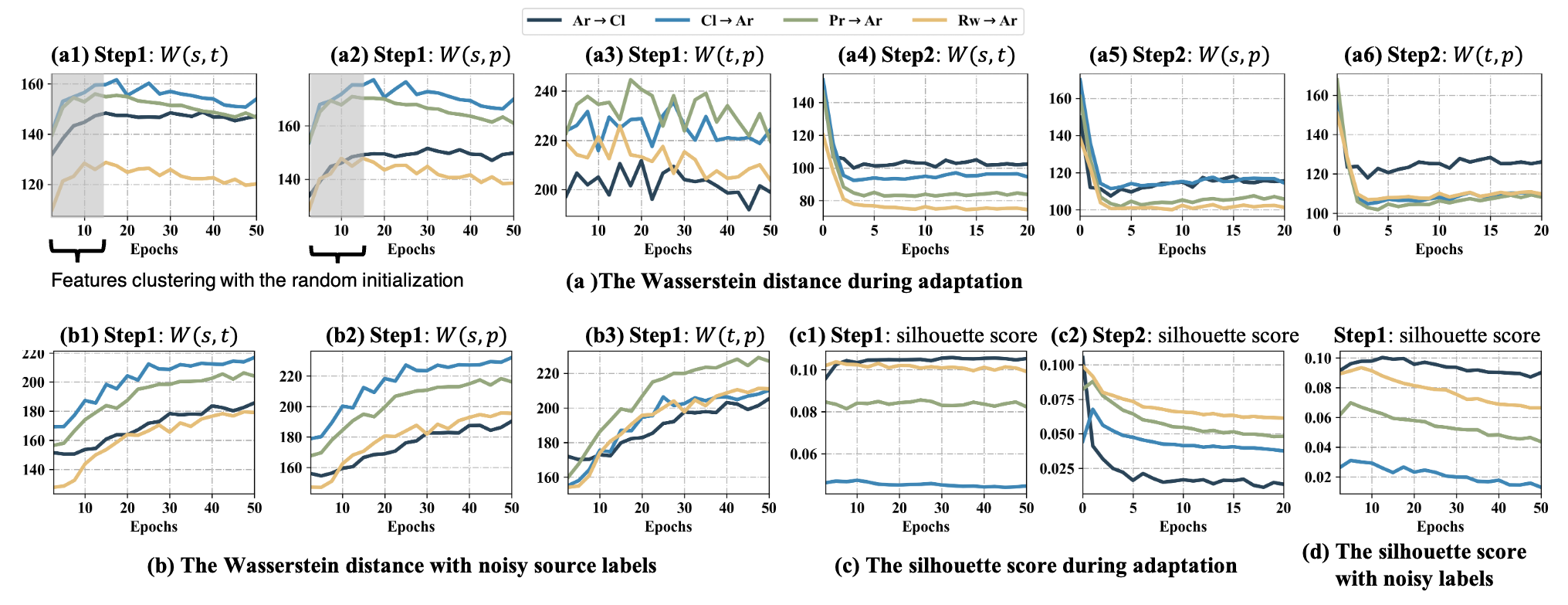}
    \caption{Empirical analysis of the behaviour of source domain, target domain and pre-training data on Office-Home\cite{venkateswara2017deep}. (a) \textbf{Pre-training Gravity.} The distribution discrepancies between pre-training data and source/target domains decrease during adaptation. (b) The increased distribution discrepancy shows that pre-training gravity can be weakened or reversed by noisy labels. (c-d) \textbf{Pre-trained Knowledge Degeneration.} The silhouette scores of pre-trained data remain stable with labelled source data in Step 1. The decline in step 2 and when training with noisy labels shows that pre-trained semantic knowledge undergoes degeneration exclusively when dealing with the noisy labels during adaptation.}
    \label{fig:W-dis}
\end{figure*}

\section{Analysis of the Impact of Pre-training on Adaptation}


\noindent \textbf{Experiment Setup:} Let $\mathcal{D}_p=\{(\textbf{x}_p,\textbf{y}_p)\}$ denote the pre-training data drawn from $P_p$, and $\mathcal{L}_p$ denote the pre-training objective. For all the analysed results in this section, we adopt the SFUDA setup SHOT~\cite{liang2020we} on Office-Home~\cite{venkateswara2017deep} as the baseline method. \revise{Following SHOT \cite{liang2020we}, the network consists of a feature extractor, comprising a ResNet-50 backbone followed by a linear bottleneck layer, and a classifier.} It proposes the following two steps as illustrated in \cref{fig:enter-label}: \textit{in step 1}, we fine-tune the ImageNet~\cite{deng2009imagenet} pre-trained model $\theta_p$ using labelled source data, resulting model $\theta_s$; \textit{in step 2,} we fine-tune model $\theta_s$ using target data with pseudo-labels.

\noindent \textbf{Metric:} We dynamically analyse two metrics during adaptation to explore the behaviour of three domains.
\begin{itemize}
    \item \textit{Wasserstein distance}: we utilize Wasserstein distance, denoted as $W(\cdot, \cdot)$, between two domains' features (any two among source $s$, target $t$, and pre-training data $p$ domain) to measure the domain discrepancy~\cite{shen2018wasserstein, kono2021wasserstein}.
   \item \textit{Silhouette score}: As features from the same class tend to form clusters in the feature space~\cite{kilinc2018learning, hess2020softmax}, we utilize the silhouette score, which quantifies intra-cluster consistency, to evaluate the pre-trained knowledge.
\end{itemize}

\subsection{Does Pre-training Impact the Adaptation?}
\label{pg}

\begin{Finding} \textbf{(Pre-training Gravity) }
The pre-training data exhibit attraction to the source and target domains in the feature space, showing the impact of pre-training during adaptation.
\end{Finding}

Intuitively, the source and target domains are aligned during adaptation \cite{zhang2019bridging, Li21BCDM} and it tends to result in the decreased discrepancy $W(s, t)$. Interestingly, in \cref{fig:W-dis}(a), we observe that the following \textit{two discrepancies continuously decrease as well}, \ie, the one between pre-training and source $W(s, p)$ and the other between pre-training and target $W(t, p)$.  We term the phenomenon of pulling source and target features closer to the pre-training feature space in adaptation as the \textbf{\textit{Pre-training Gravity}}. Note that the above two \textit{discrepancies decreased simultaneously in both step 1 and step 2}, even though the target domain is \textit{not} involved in step 1 (source-only fine-tuning), and the source domain is \textit{not} involved in step 2 (target-only fine-tuning). We hypothesize the reason to be the inherent similarity between the source and target domains.

Moreover, it is observed that the Wasserstein distance is stable and then slightly increases at the final epochs of step 2 (\cref{fig:W-dis}(a4-a6)). We conjecture that the pseudo-label noise in step 2 controls the Pre-training Gravity. To verify it, we manually generate noisy labels by replacing 50\% of the ground-truth labels with random labels and fine-tune the model with these noisy labels. The results are shown in \cref{fig:W-dis}(b). The significant increase of three discrepancies suggests that pre-training gravity can be weakened or destroyed by noisy labels because they conflict with the pre-trained knowledge. Similarly, we conjecture that the increased discrepancy in step 2 is caused by the noisy target pseudo-labels. 

\revise{The Wasserstein distance exhibits different behaviors in the two steps. First, as shown in \cref{fig:W-dis}(a1-a2), there is an initial increase during the early epochs of step 1. This occurs because the bottleneck and classifier are randomly initialized, leading to extracted features that lack meaningful representation and are clustered together. Consequently, the Wasserstein distance remains small at the beginning. As learning progresses, the model extracts more discriminative features, which contains the domain discrepancy, resulting in a rise in the Wasserstein distance. Second, the differing rates of descent in the two stages stem from variations in knowledge transfer speed, influenced by model initialization and training data. Specifically, as shown in Fig. 3, in step 1 (\cref{fig:W-dis}(a1-a2)), the model is initialized from the pre-trained model and trained on source data. In step 2 (\cref{fig:W-dis}(a4-a6)), the model is initialized from the source-trained model and trained on target data. Since the domain discrepancy between pre-training and source data is greater than that between source and target data—illustrated by $W(s, p)$ being larger than $W(s, t)$ in \cref{fig:W-dis}(a)—knowledge transfer occurs more rapidly in step 1 than in step 2.}

\begin{Discussion}
\textbf{\textit{Pre-training Gravity}} elucidates from another view why SFUDA methods \cite{liang2020we, pei2023uncertainty} are still effective when source and target data is not available simultaneously: two domains are implicitly aligned, as illustrated in \cref{teaser}(a). 
\end{Discussion}

\subsection{How Does the Pre-training Impact the Target Error?}  
\label{how}
\begin{Factor} \textbf{(Pre-trained Knowledge Degeneration)}
Pre-trained semantic knowledge undergoes degeneration exclusively when dealing with the noisy target pseudo-labels during adaptation. Target errors also come from the attraction of the degenerated pre-trained semantic knowledge due to the noisy pseudo-labels.
\end{Factor}

The stable silhouette scores in step 1 (\cref{fig:W-dis}(c)) demonstrate that pre-trained knowledge is preserved during fine-tuning on labelled source data. In contrast, the decreased scores during adaptation to the target domain in step 2 suggest the degradation of the pre-trained knowledge (\cref{fig:W-dis}(c)). We hypothesize that the cause of the above pre-trained knowledge degeneration is the noise in the target pseudo labels. To verify it, we train the model with 50\% of the ground-truth replaced with random labels. It is observed in \cref{fig:W-dis}(d) that silhouette scores also decrease, indicating that pre-trained semantic knowledge degenerates when dealing with noisy labels. We term this phenomenon as \textit{Pre-trained Knowledge Degeneration} (See \cref{tsne_car} for more visualization). 

\begin{Discussion} 
\textbf{\textit{Pre-trained knowledge degeneration}} fundamentally differs from catastrophic forgetting. While catastrophic forgetting \cite{kirkpatrick2017overcoming} refers to a model's tendency to forget previously learned knowledge when exposed to new data or tasks, pre-trained knowledge degeneration occurs exclusively in the presence of noisy labels and does not occur with ground truth. The degradation stems from the conflict between the noisy labels and the pre-learned knowledge in the model. During adaptation, \textbf{\textit{Pre-training Gravity}} is weakened but still exists due to the small amount of noisy target pseudo-labels. Meanwhile, the pre-trained knowledge undergoes degeneration, concurrently attracting the target domain and thereby increasing the target error.
\end{Discussion}

\begin{Factor} \textbf{(Gradient Disparity)}
Target error depends on the gradient disparity between pre-training data and target domain.
\end{Factor}
We present a theoretical analysis of the impact of pre-training through the lens of stochastic gradient descent and generalization. We first define the target excess risk to quantify the model's performance on the target domain, and gradient disparity to describe the approximation accuracy when replacing fine-tuning on target data with other domains and objective functions.

\begin{figure*}[t]
    \centering
    \includegraphics[width=\linewidth]{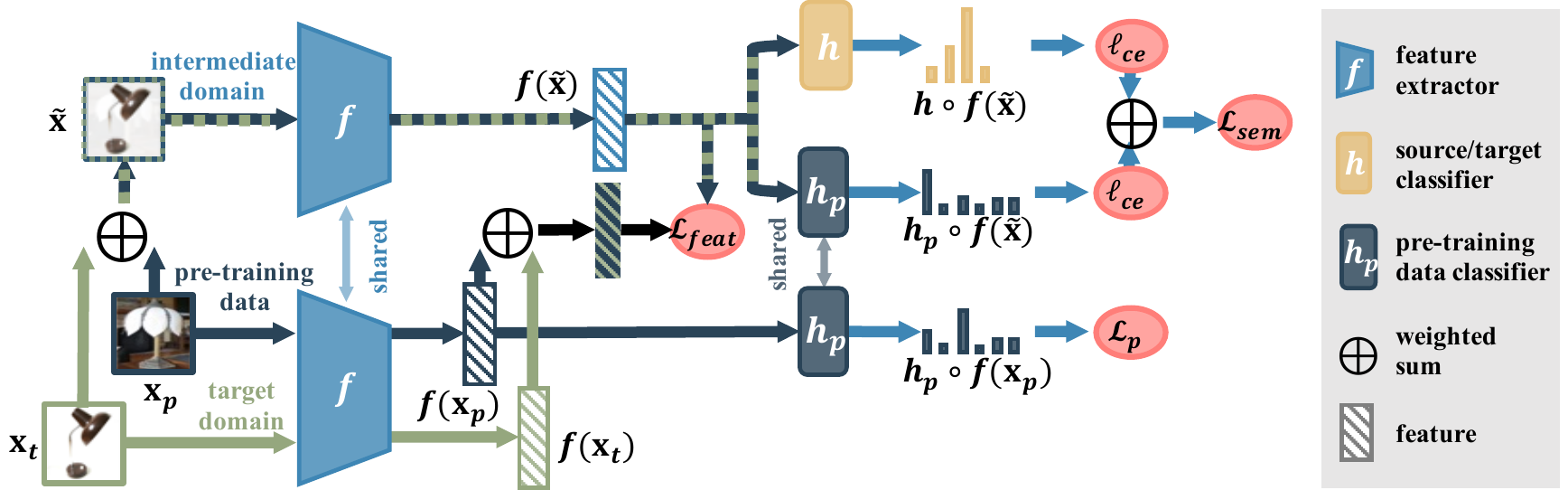}
    \caption{Training pipeline of TriDA. We make minimal architecture changes and only added a lightweight classification $h_p(\cdot)$ to the existing method. Thus, TriDA can seamlessly complement existing UDA and SFUDA methods. Only the target domain and pre-training data are depicted for brevity. TriDA reuses pre-training data with $\mathcal{L}_p$ to mitigate pre-trained knowledge degeneration and to reduce the impact of gradient disparity. In addition, the intermediate domain between target domain and pre-training data with $\mathcal{L}_{sem}$ and $\mathcal{L}_{feat}$ to reduce the impact of gradient disparity to lower the performance boundary.}
    \label{pipeline}
\end{figure*}

\begin{Definition} \textbf{(Target Excess Risk)}
Let $\mathcal{R}_t(\theta)$ be the oracle cross-entropy for the model $\theta$ on the target domain. We define the target excess risk $\widetilde{\mathcal{R}}_t(\theta)$ as:
\begin{align}
\widetilde{\mathcal{R}}_t(\theta) = \mathcal{R}_t(\theta) - \mathcal{R}_t(\theta^\ast), \text{\ where \ } \theta^\ast = \arg\min_{\theta} \mathcal{R}_t(\theta).
\end{align}
\end{Definition}

\begin{Definition} \textbf{(Gradient Disparity)}
Let $\mathcal{L}(\theta)$ be an arbitrary proposed objective function that the model aims to optimize, where $\theta$ represents the model parameters and $\nabla$ denotes the gradient operator with respect to $\theta$. We define the gradient disparity $\Delta({\mathcal{L}})$, which is
the least upper bound of the absolute value of difference between $\nabla \mathcal{L}(\theta)$ and $\nabla \mathcal{R}_t(\theta)$:
\begin{align}
\Vert \nabla \mathcal{L}(\theta) - \nabla \mathcal{R}_t(\theta)\Vert \leq \Delta(\mathcal{L}), \forall \theta
\end{align}
\end{Definition}

From the perspective of learning, the difference of two gradients quantifies the information difference between two domains \cite{peng2022learning}. Then, we present the dependency of target excess risk on the gradient disparity in UDA and SFUDA scenarios, respectively (see supplementary material for proof).


\begin{Proposition} \textbf{(Pre-training and Source Fine-tuning)}
Starting from a random initialization $\theta_0$, the model is trained to optimize $\mathcal{L}_{p}(\theta)$ on pre-training data, resulting in the pre-trained model $\theta_p$. Subsequently, the pre-trained model is fine-tuned on the source domain with the objective function $\mathcal{L}_{s}(\theta)$ for $T_s$ iterations, resulting in the source fine-tuned model $\theta_s$. We have the following performance guarantees for model $\theta_p$ and $\theta_s$:
\begin{align}
&\widetilde{\mathcal{R}}_t(\theta_p) \leq \mathcal{O}(\Delta(\mathcal{L}_p)^2),\\
&\widetilde{\mathcal{R}}_t(\theta_s) 
\leq \mathcal{O}(\log(T_s \Delta(\mathcal{L}_p)^2)/T_s + {\Delta(\mathcal{L}_s)^2}).
\end{align}
\end{Proposition}



\begin{Proposition} \textbf{(Vanilla UDA)}
Starting from pre-trained initialization $\theta_p$, the model is trained with the objective function $\mathcal{L}_{\gamma}(\theta) = \mathcal{L}_s(\theta)+\mathcal{L}_{UDA}(\theta)$, which combines the source-only loss $\mathcal{L}_s(\theta)$ and the adaptation loss $\mathcal{L}_{UDA}(\theta)$ jointly on source and target domains. The adapted model $\theta_t$ is obtained after $T_\gamma$ iterations. We have the performance guarantee:
\begin{align}
\label{eqvuda}
&\widetilde{\mathcal{R}}_t(\theta_t) \leq \mathcal{O}(\log(T_\gamma \Delta(\mathcal{L}_p)^2)/T_\gamma + {\Delta(\mathcal{L}_\gamma)^2}).
\end{align}
\end{Proposition}


\begin{Proposition} \textbf{(SFUDA)}
Starting from the source fine-tuned initialization $\theta_s$, the model is trained on the unlabelled target domain with objective function $\mathcal{L}_{SFUDA}(\theta)$. The adapted model $\theta_t$ is obtained after $T_t$ iterations. We have the performance guarantee:
\begin{align}
\label{eqsfuda}
\widetilde{\mathcal{R}}_t(\theta_t) \leq \mathcal{O}(\log(T_t \Delta(\mathcal{L}_p)^2)/T_t + \exp(-T_t) {\Delta(\mathcal{L}_s)^2} \notag \\  + {\Delta(\mathcal{L}_{SFUDA})^2}).
\end{align}
\end{Proposition}



 
\begin{Discussion}
Theoretical analysis indicates that the performance of adaptation is closely related to the gradient disparities, and we categorize them into two components: (1) $\Delta(\mathcal{L}_p)$, which represents the gradient disparity caused by pre-training data and target data; and (2) $\Delta(\mathcal{L}_s)$, $\Delta(\mathcal{L}_\gamma)$, and $\Delta(\mathcal{L}_t)$, which represents the gradient disparity caused by source and target data. While two-domain paradigm methods address the second component by aligning source and target domains and pseudo-label algorithms, they overlook $\Delta(\mathcal{L}_p)$. This oversight becomes particularly noticeable in scenarios with limited source and target samples, as the number of iterations ($T$) correlates with the dataset size.  
\end{Discussion}

\subsection{Could We Improve Target Error Bound via Incorporating Pre-training Data?}

The above findings motivate us to incorporate pre-training data into the adaptation process. First, we can address pre-trained knowledge degeneration by directly imposing constraints on the pre-training data. We then demonstrate the impact of these constraints on the target error bound.

\begin{Proposition} \label{Pro:incorporate} \textbf{(Incorporating Pre-training Data)}
Suppose we incorporate pre-training data into training using the objective function $\Phi(\theta):=\mathbb{E}_{(\textbf{x}_p, \textbf{y}_p)\sim P_p}\phi(\theta;\textbf{x}_p, \textbf{y}_p)$. Importantly, this objective function need not be consistent with the pre-training objective function $\mathcal{L}_p$. We combine it with the adaptation objective function  $\mathcal{L}_{\gamma}(\theta)$, weighted by $\alpha\in(0,1)$, resulting in $\alpha \mathcal{L}_{\gamma}(\theta) + (1-\alpha)\Phi(\theta)$. We have a performance guarantee for adapted model $\theta_t$:
\begin{align}
\label{modif}
\widetilde{\mathcal{R}}_t(\theta_t) \leq \mathcal{O}(\alpha \log(T_\gamma \Delta(\mathcal{L}_p)^2 / \alpha) /T_\gamma + \alpha{\Delta(\mathcal{L}_\gamma)^2} \notag \\ + (1-\alpha) \delta^2),\notag \\
\delta^2 = \max_{\theta,(\textbf{x}_p,\textbf{y}_p)}[\mathbb{E}[\Vert \nabla \phi(\theta;\textbf{x}_p, \textbf{y}_p) - \nabla \mathcal{R}_t(\theta) \Vert ]].
\end{align}
\end{Proposition}

\begin{Discussion}
When combining $\Phi(\theta)$ on the pre-training data, the impact of pre-training is reduced, because the term $\log(T_\gamma \Delta(\mathcal{L}_p)^2)/T_\gamma$ in \cref{eqvuda} is reduced to $\alpha \log(T_\gamma \Delta(\mathcal{L}_p)^2 / \alpha) /T_\gamma$ where $\alpha\in(0,1)$. Consequently, the bound of $\widetilde{\mathcal{R}}_t(\theta_t)$ includes an additional term, $\delta^2$. A smaller $\delta^2$ allows the target excess risk $\widetilde{\mathcal{R}}_t(\theta_t)$ to have a lower bound compared to \cref{eqvuda}. Significantly, $\delta^2$ is determined by the pre-training data $(\textbf{x}_p, \textbf{y}_p)$ and is associated with the objective function $\phi(\cdot)$. \textit{This finding encourages us to reduce the target error bound by selectively incorporating pre-training data and carefully designing the objective function for the pre-training phase.}
\end{Discussion}


\section{Method: The Three-Domain Problem}
\label{method}
Inspired by the analysis above, we extend the two-domain paradigm by incorporating pre-training data into adaptation and redefining UDA as a three-domain problem: source, target, and pre-training data domains. We introduce a novel framework termed TriDA to address the target error stemming from the pre-training \ie pre-trained knowledge degeneration and 
the target risk boundary in \cref{modif}. Notably, TriDA can seamlessly complement existing vanilla UDA/SFUDA methods with objective $\mathcal{L}_{UDA}$/ $\mathcal{L}_{SFUDA}$. 

In the following, we first give the architecture and model optimization in \cref{framework}. After that, we provide the selection strategy for pre-training data in \cref{sec:select}, followed by the solution when the pre-training data is unavailable in \cref{Inaccessible}.

\subsection{The \mName Framework}
\label{framework}
\textbf{Architecture.} As illustrated in \cref{pipeline}, the network consists of a feature extractor $f(\cdot)$, a classifier $h(\cdot)$ for the source and target domains, and a classifier $h_p(\cdot)$ for pre-training data. We make minimal architecture changes and only added a lightweight classification $h_p(\cdot)$ to the existing method, which can be removed after training, thereby enhancing the adaptation performance \textit{without} additional inference costs.

\noindent \textbf{Mitigate Pre-trained Knowledge Degeneration.}
We propose a straightforward solution by introducing cross-entropy loss, denoted as $\ell_{ce}$, for the raw pre-training images. This approach allows manual intervention to preserve the pre-trained knowledge. The objective function is as follows (we omit $\theta$ for brevity):
\begin{align}
    \mathcal{L}_{p} = \ell_{ce}(h_p\circ f(\textbf{x}_p),\textbf{y}_p).
\end{align}

\noindent \textbf{Lower the Target Risk Boundary.} \cref{Pro:incorporate} shows that incorporating pre-training data reduces its impact but introduces the term $\delta^2$. Our goal is to achieve a lower target excess risk boundary by minimizing $\delta^2$. This term is influenced by two key factors: the objective function $\phi(\theta;\cdot)$ and the used pre-training data $(\textbf{x}_p,\textbf{y}_p)$. In this section, we explore the design of $\Phi(\theta)$. (The selection of pre-training data is discussed in \cref{sec:select}). 

We introduce an intermediate domain that incorporates information from both the pre-training data and the target domain. We show that applying the objective function to this intermediate domain helps achieve a lower $\delta^2$. Specifically, the intermediate domain data $\widetilde{\textbf{x}}$ is generated by the weighted sum: $\widetilde{\textbf{x}}=\lambda \textbf{x}_p + (1-\lambda) \textbf{x}_t$, where $\lambda \sim Beta(\alpha, \alpha)$.

\begin{Proposition} \label{incordelt}\textbf{(Incorporating Intermediate Domain Data)} 
Suppose we use cross-entropy as the objective function $\phi(\theta)$ on pre-training data. Let $\hat{\textbf{y}}_t$ be the target pseudo-label.  $\delta^2$ in \cref{modif} is transformed to $\widetilde{\delta}^2$, and we have:
\begin{align}
\label{widedelta}
    \widetilde{\delta}^2 &= \max_{\theta,(\textbf{x}_p,\textbf{y}_p)}[\mathbb{E}[\Vert \nabla \phi(\theta;\widetilde{\textbf{x}}, \textbf{y}_p, \hat{\textbf{y}}_t) - \nabla \mathcal{R}_t(\theta) \Vert \notag], \\ \widetilde{\delta}^2 &\leq \delta^2 = \max_{\theta,(\textbf{x}_p,\textbf{y}_p)}[\mathbb{E}[\Vert \nabla \phi(\theta;\textbf{x}_p, \textbf{y}_p) - \nabla \mathcal{R}_t(\theta) \Vert ]].
\end{align}
\end{Proposition}
\cref{incordelt} demonstrates that proposing an objective function on the intermediate domain achieves $\widetilde{\delta}^2 \leq \delta^2$, resulting in a lower risk boundary than Proposition \ref{Pro:incorporate}, which directly applying on the source domain in  (see supplementary material for proof). The generation of pseudo-labels is not the primary focus of this paper, we follow the pseudo-label generation in baselines \cite{liang2020we,lee2022confidence}. Since the semantic spaces of the two domains do not overlap, we supervise the predictions of $h(\cdot)$ and $h_p(\cdot)$ with labels from the respective domains:
\begin{align}
    \mathcal{L}_{sem} = &\lambda\ell_{ce}(h_p\circ f(\widetilde{\textbf{x}}),\textbf{y}_p) \notag\\ &+  (1-\lambda)\ell_{ce} (h\circ f(\widetilde{\textbf{x}}),\hat{\textbf{y}}_t).
\end{align}

Unlike mixup~\cite{zhang2018mixup} which mixes intra-domain data and labels as a regularization to enhance generalization and mitigate overfitting \cite{zhang2020does}, we involve mixing inter-domain data (pre-training data and target data) from distinct label spaces. We then impose constraints on two classifiers to minimize the gradient disparity. Furthermore, we introduce the regularization in the feature space \cite{verma2019manifold} to bridge the features of the two domains by enhancing local smoothness:
\begin{equation}
\begin{aligned}
    \mathcal{L}_{feat} &= \Vert\lambda f(\textbf{x}_p) + (1-\lambda) f(\textbf{x}_t)
 - f(\widetilde{\textbf{x}})\Vert_1.
\end{aligned}
\end{equation}


\noindent \textbf{Overall objective.}  Our method is plug-and-play. Let $\beta$ denote the trade-off parameter. For Vanilla UDA methods with objective $\mathcal{L}_{UDA}$, the overall objective is:
\begin{align}
    \min_{f,h,h_p} \mathcal{L}_{UDA} + \mathcal{L}_s + \mathcal{L}_p  + \beta( \mathcal{L}_{sem} + \mathcal{L}_{feat}).
\end{align}

For SFUDA, in step 1, we aim to train a model on the labelled source domain to serve as the initialization for target adaptation. We adhere to the standard practice of minimizing the source error: $\mathcal{L}_s = \ell_{ce} (h\circ f(\textbf{x}_s),\textbf{y}_s)$. Thus, the objective is:
\begin{align}
\centering
    \min_{f,h,h_p} \mathcal{L}_{s} + \mathcal{L}_{p}.
\end{align}
In step 2, for SFUDA methods with objective $\mathcal{L}_{SFUDA}$, the overall objective is:
\begin{align}
\label{betae}
    \min_{f,h,h_p} \mathcal{L}_{SFUDA} + \mathcal{L}_{p} + \beta( \mathcal{L}_{sem} + \mathcal{L}_{feat}).
\end{align}

In summary, pre-trained knowledge degeneration is mitigated by $\mathcal{L}_{p}$, and the target risk boundary is decreased by $\mathcal{L}_{sem}$ and $\mathcal{L}_{feat}$. Thus, TriDA effectively reduces the error stemming from pre-training data.



\subsection{Pre-training Data Selection}
\label{sec:select}

As $\widetilde{\delta}^2$ in \cref{widedelta} is related to pre-training data $(\textbf{x}_p,\textbf{y}_p)$, we introduce a class-based selection strategy on the utilized pre-training data to further minimize $\delta^2$ for a lower target risk boundary. In addition, the pre-training dataset is significantly larger than existing UDA benchmarks. The data selection substantially reduces the volume of pre-training data, thereby enhancing efficiency. 

Notably, not all categories presented in the target domain have counterparts in the pre-training dataset due to different annotation granularity. Thus, we select categories with high semantic similarity to the target classes, as they are more likely to exhibit greater proximity in the feature space, as illustrated in \cref{tsne_car}. Let $\mathcal{C}_p=\{c_{p,(i)}\}$ and $\mathcal{C}_t=\{c_{t,(i)}\}$ be the classes in the pre-training dataset and target domain, respectively. Let $\text{sim}(\cdot, \cdot)$ be the semantic similarity based on the shortest path connecting the synset in the WordNet\cite{miller1998wordnet} hierarchy tree structure. For instance, ImageNet does not include the category \texttt{"Telephone"} from Office-Home, and we find that the top-2 similar classes are \texttt{"Pay-phone"} and \texttt{"Dial telephone"}. Therefore, we select pre-training classes whose maximum similarity to a target class exceeds a threshold $\tau$:
\begin{align}
\{c_{p,(i)}| \max_j [\text{sim}(c_{p,(i)}, c_{t,(j)})] > \tau \}.
\end{align}

\subsection{When Pre-training Data is Unavailable}
\label{Inaccessible}
In \cref{framework}, we presented a solution when pre-training data is accessible. However, the raw data may be unavailable due to data privacy, security, and proprietary restrictions. To address it, we provide an alternative solution for scenarios where only the pre-trained model is available. We aim to synthesize proxy pre-training images with the knowledge of the model weight. Specifically, given the label $\textbf{y}_i$ for class $c_{p,(i)}$, we keep the model weight fixed and optimize $\hat{\textbf{x}}_i$ from a random initialization image with the objective:
\begin{align}
\label{synobj}
\min_{\hat{\textbf{x}}_i} \ell_{ce}(h_p \circ f(\hat{\textbf{x}}_i), \textbf{y}_i) + Reg(\hat{\textbf{x}}_i).
\end{align}
Following \cite{yin2020dreaming} $Reg(\cdot)$ is the regularization consisting of the total variance, $\ell_2$ norm, and feature regularization components (see supplementary material for details).

\begin{table}
\small
\centering
\caption{Class selection settings. Denote $N_c$ by the number of selected classes.}
\begin{tabular}{lcc}
\toprule
Dataset& $\tau$& $N_c$\\
\midrule
Office31& 0.2& 72\\
Office-Home& 0.4& 63\\
VisDA-C& 0.2& 59\\
\bottomrule
\end{tabular}
\label{tab:class}
\end{table}

\begin{table*}[h]
    \centering
     \caption{Classification accuracy(\%) on Office-31 (ResNet50). Bold and underlined numbers denote the top-1 and second best accuracy, respectively.}
    \begin{tabular*}{0.8 \hsize}{@{}@{\extracolsep{\fill}}clccccccc@{}}
    \toprule
    Task& Method  & A$\rightarrow$D& A$\rightarrow$W& D$\rightarrow$A& D$\rightarrow$W& W$\rightarrow$A& W$\rightarrow$D& Avg. \\
    \midrule
    \multirow{5}{*}{\rotatebox{90}{UDA}}
    &ResNet-50 \cite{he2016deep}& 79.3& 75.8& 63.6& 95.5& 63.8& 99.0& 79.5\\
    &MCD \cite{saito2018maximum}& 87.3& 90.4& 68.3& {98.5}& 67.6& \textbf{100.0}& 85.4\\
    \cline{2-9}
    &CDAN \cite{long2018conditional}&  89.9& 93.8& 73.4& {98.5}& 70.4& \textbf{100.0}& 87.7\\
    &TriDA+CDAN& {92.2}& {95.1}& {75.9}& {98.5}& \underline{74.6}& \textbf{100.0}& {89.4}\\
    &MDD \cite{zhang2019bridging}& \underline{94.4}& \underline{95.6}& \underline{76.6}& \underline{98.6}& {72.2}& \textbf{100.0}& \underline{89.6}\\
    &\revise{TriDA+MDD} &\revise{\textbf{95.2}} &\revise{\textbf{96.1}} &\revise{\textbf{78.3}}& \revise{\textbf{98.7}}&\revise{\textbf{75.6}}&\revise{\textbf{100.0}}&\revise{\textbf{90.7}}\\
    \midrule
    \midrule
    \multirow{11}{*}{\rotatebox{90}{SFUDA}}
    &SFDA \cite{kim2021domain}&  92.2& 91.1& 71.0& 98.2& 71.2& 99.5& 87.2\\
    &A$^2$Net \cite{xia2021adaptive}&   94.5& 94.0& \underline{76.7}& \textbf{99.2}& 76.1& \textbf{100.0}& 90.1\\
    &NRC \cite{yang2021exploiting} &  96.0& 90.8& 75.3&  99.0& 75.0& \textbf{100.0}&  89.4\\
    &HCL \cite{huang2021model}&  94.7& 92.5& 75.9& 98.2& \textbf{77.7}& \textbf{100.0}&  89.8\\
    &DIPE \cite{wang2022exploring}&  \textbf{96.6}& 93.1& 75.5& 98.4& 77.2& 99.6& 90.1\\
    &AaD \cite{yang2022attracting}&  \underline{96.4}& 92.1& 75.0& \underline{99.1}& 76.5& \textbf{100.0}& 89.9\\
    &SFDA-DE \cite{ding2022source}& 96.0& 94.2& 76.6& 98.5& 75.5& 99.8& 90.1\\
    \cline{2-9}
    &CoWA\cite{lee2022confidence}& 94.4& \underline{95.2}& 76.2& 98.5& \underline{77.6}& 99.8& \underline{90.3}\\
    &TriDA+CoWA &93.6 &\textbf{96.2} &\textbf{77.1} &98.6 &76.3 &\textbf{100.0} &\underline{90.3}\\
    &SHOT \cite{liang2020we}& 94.0& 90.1& 74.7& 98.4& 74.3& \underline{99.9}& 88.6\\
    &TriDA+SHOT &94.8 &94.6 &\textbf{77.1} &98.9 &76.8 &\textbf{100.0} &\textbf{90.4}\\
    \bottomrule
\end{tabular*}
\vspace{-3mm}
    \label{tab:office31}
\end{table*}

Moreover, we extend our methods to the multi-modal pre-trained model CLIP \cite{Radford2021LearningTV} whose pre-training dataset is unavailable. As CLIP is trained on language-image pairs with contrastive learning, and the label is a sentence describing the image, we cannot directly regard the sentences as classes and select from them. Addressing this, we synthesize images corresponding to the classes of the target domain using the prompt “a photo of \texttt{[CLASS]}”. Let $w(c_{t,(i)})$ be the text embedding for target class $c_{t,(i)}$. We replace the classification probability from $h_p \circ f(\hat{\textbf{x}}_i)$ in \cref{synobj} with:
\begin{align}
\frac{\exp(\text{sim}(f(\hat{\textbf{x}}_i), w(c_{t,(i)}))/t)}{\sum_{c_{t,(j)}\in\mathcal{C}_t}\exp(\text{sim}(f(\hat{\textbf{x}}_j), w(c_{t,(j)}))/t)},
\end{align}
where $t$ is temperature learned by CLIP.

\begin{table*}
\small
\caption{Classification accuracy (\%) on Office-Home (ResNet50). Bold and underlined numbers denote the top-1 and second best accuracy, respectively.}
\vspace{-6mm}
\label{tab:officehome}
\begin{center}
\resizebox{\linewidth}{!}{
\begin{tabular}{clccccccccccccccc}
\toprule
Task& Method  & Ar$\rightarrow$Cl&	Ar$\rightarrow$Pr& Ar$\rightarrow$Rw& Cl$\rightarrow$Ar&	Cl$\rightarrow$Pr& Cl$\rightarrow$Rw& Pr$\rightarrow$Ar& Pr$\rightarrow$Cl& Pr$\rightarrow$Rw&	Rw$\rightarrow$Ar& Rw$\rightarrow$Cl& Rw$\rightarrow$Pr& Avg.\\
\midrule
\multirow{5}{*}{\rotatebox{90}{UDA}}
&ResNet-50 \cite{he2016deep}&   41.1& 65.9& 73.7& 53.1& 60.1& 63.3& 52.2& 36.7& 71.8&     64.8& 42.6& 75.2& 58.4\\
&MCD \cite{saito2018maximum}&	51.7& 72.2& \underline{78.2}& \underline{63.7}& 69.5& 70.8& 61.5& 52.8& 78.0& \underline{74.5}& 58.4& 81.8& 67.8\\

\cline{2-15}
&CDAN \cite{long2018conditional}& {55.2}& 72.4& 77.6& 62.0& 69.7& 70.9& 62.4& 54.3& {80.5}& \textbf{75.5}& {61.0}& 83.8& 68.8\\
&TriDA+CDAN& 54.7& {74.2}& \underline{79.6}& \textbf{66.8}& \textbf{75.4}& \underline{76.1}& \textbf{66.4}& \underline{54.9}& \textbf{82.4}& 74.4& \textbf{62.2}& {84.0}& \underline{70.9}\\
&MDD \cite{zhang2019bridging}&  \underline{56.2}& \underline{75.4}& \underline{79.6}& 63.5& {72.1}& {73.8} & {62.5}& {54.8}& 79.9& 73.5& 60.9& \textbf{84.5}& {69.7}\\
&\revise{TriDA+MDD}& \revise{\textbf{57.5}}& \revise{\textbf{76.8}}& \revise{\textbf{80.8}}& \revise{\underline{66.5}}& \revise{\underline{75.2}}& \revise{\textbf{77.4}}& \revise{\underline{65.3}}& \revise{\textbf{55.3}} &\revise{\underline{81.1}}& \revise{73.7}& \revise{\textbf{62.2}}& \revise{\textbf{84.5}} & \revise{\textbf{71.4}}\\
\midrule
\midrule

\multirow{11}{*}{\rotatebox{90}{SFUDA}}

&SFDA \cite{kim2021domain}&  48.4& 73.4& 76.9& 64.3& 69.8& 71.7& 62.7& 45.3& 76.6& 69.8& 50.5& 79.0& 65.7\\
&A$^2$Net \cite{xia2021adaptive}&  58.4& 79.0& \underline{82.4}& 67.5& 79.3& 78.9& 68.0& 56.2& 82.9& 74.1& 60.5&  {85.0}& 72.8\\
&NRC \cite{yang2021exploiting}& 57.7& \textbf{80.3}& 82.0& 68.1& 79.8& 78.6& 65.3& 56.4& 83.0& 71.0& 58.6& \underline{85.6}& 72.2\\
&DIPE \cite{wang2022exploring}& 56.5& 79.2& 80.7& \textbf{70.1}& 79.8& 78.8& 67.9& 55.1& \underline{83.5}& \underline{74.1}& 59.3& 84.8& 72.5\\
&AaD \cite{yang2022attracting}& \underline{59.3}& 79.3& 82.1& 68.9& 79.8& 79.5& 67.2& \textbf{57.4}& 83.1& 72.1& 58.5& 85.4& 72.7\\
&SFDA-DE \cite{ding2022source}& \textbf{59.7}& {79.5}& \underline{82.4}& \underline{69.7}& 78.6& 79.2& 66.1&  \underline{57.2}& 82.6& 73.9& \underline{60.8}& \underline{85.5}& {72.9}\\

&DaC \cite{zhang2022divide}& 59.1& {79.5}& 81.2& 69.3& 78.9& 79.2& 67.4& 56.4& 82.4& 74.0& \textbf{61.4}& 84.4& 72.8 \\
 \cline{2-15}
 &CoWA \cite{lee2022confidence}&  56.9& 78.4& 81.0& 69.1& 80.0& 79.9& 67.7&  \underline{57.2}& 82.4& 72.8& 60.5& 84.5& 72.5\\
&TriDA+CoWA& 58.7& 77.7& 82.2& 69.6&\textbf{82.0}& \underline{81.7}& \underline{69.3}& 56.2& \textbf{83.7}& 73.8& 60.5& 84.9& \underline{73.4}\\
 &SHOT \cite{liang2020we}&   57.1& 78.1& 81.5& 68.0& 78.2& 78.1& 67.4& 54.9& 82.2& 73.3& 58.8& 84.3& 71.8\\
&TriDA+SHOT& 57.3& \underline{80.2}& \textbf{83.1}& \textbf{70.1}& \underline{81.3}& \textbf{82.1}& \textbf{70.7}& 54.7& 83.0& \textbf{74.3}& 59.8& \textbf{85.7}& \textbf{73.5}\\




\bottomrule
\end{tabular}}
\vspace{-6mm}
\end{center}

\end{table*}






\begin{table*}[t]
\small
\caption{Classification accuracy (\%) on Office-Home (ViT). Bold and underlined numbers denote the top-1 and second best accuracy, respectively.}
\label{tab:officehome_vit}
\begin{center}
\resizebox{\linewidth}{!}{
\begin{tabular}{clccccccccccccccc}
\toprule
Task& Method  & Ar$\rightarrow$Cl&	Ar$\rightarrow$Pr& Ar$\rightarrow$Rw& Cl$\rightarrow$Ar&	Cl$\rightarrow$Pr& Cl$\rightarrow$Rw& Pr$\rightarrow$Ar& Pr$\rightarrow$Cl& Pr$\rightarrow$Rw&	Rw$\rightarrow$Ar& Rw$\rightarrow$Cl& Rw$\rightarrow$Pr& Avg.\\
\midrule
\multirow{5}{*}{\rotatebox{90}{UDA}}
&ViT \cite{dosovitskiy2020image}&52.4 &82.1 &86.9 &76.8 &84.1	&86.0 &75.1 &51.2 &88.1 &78.3 &51.5 &87.8 &75.0\\
&CDAN \cite{long2018conditional} &61.6& 87.8 &89.6 &81.4& 88.1&88.5 &82.4 &62.5 &90.8 &84.2 &63.5 &90.8 &80.9\\
&TriDA+CDAN &62.3 &87.9 &90.1 &81.3 &88.6 &89.4 &82.6 &62.5 &90.9 &84.1 &64.5 &91.4 &81.3\\
&SSRT \cite{sun2022safe} &\underline{75.2} &\underline{89.0} &\underline{91.1} &\underline{85.1} &\underline{88.3} &\underline{90.0} &\underline{85.0} &\underline{74.2} &\underline{91.3} &\underline{85.7} &\underline{78.6} &\underline{91.8} &\underline{85.4}\\
&TriDA+SSRT &\textbf{75.9} &\textbf{90.2} &\textbf{91.8} &\textbf{86.8} &\textbf{89.7} &\textbf{90.1} &\textbf{85.8} &\textbf{75.4} &\textbf{92.0} &\textbf{88.2} &\textbf{79.6} &\textbf{92.3} &\textbf{86.5}\\
\midrule
\midrule
\multirow{4}{*}{\rotatebox{90}{SFUDA}}

&CoWA \cite{lee2022confidence}&  57.0& 89.8& 90.5& \underline{83.1}& 91.3& \underline{90.3}& 81.0& 58.0& 90.4& 82.1& 57.6& 89.7& 80.1\\
&TriDA+CoWA& 59.5& \textbf{91.2}& \underline{90.8}& 83.0& \textbf{91.7}& \textbf{91.1}& \underline{82.0}& 60.8& \underline{91.6}& \underline{83.1}& 59.4& \underline{91.9}& 81.3\\
&SHOT \cite{liang2020we}&  \underline{60.6}& 89.6& 90.8& \underline{83.1}& 90.9& 90.2& 81.8& \underline{62.0}& 91.2& 83.0& \underline{61.9}& 91.1& \underline{81.4}\\
&TriDA+SHOT& \textbf{64.3}& \underline{90.9}& \textbf{91.4}& \textbf{83.6}& \underline{91.4}& \textbf{91.1}& \textbf{83.2}& \textbf{63.9}& \textbf{91.9}& \textbf{84.1}& \textbf{64.8}& \textbf{92.0}& \textbf{82.7}\\
\bottomrule
\end{tabular}}
\end{center}
\end{table*}


\begin{table*}
\small
\centering
\caption{Classification accuracy (\%) on VisDA-C (ResNet101). Bold and underlined numbers denote the top-1 and second best accuracy, respectively.}
\resizebox{\linewidth}{!}{\begin{tabular}{clcccccccccccccc}
\toprule
Task& Method& {plane}& {bike}& {bus}& {car}& {horse}& {knife}& {mcycle}& {person}& {plant}& {sktbrd}& {train}& {truck}& Avg.\\
\midrule
\multirow{5}{*}{\rotatebox{90}{UDA}}
&ResNet-101 \cite{he2016deep}& 63.6& 35.3& 50.6&  \textbf{78.2}& 74.6& 18.7& 82.1& 16.0& 84.2& 35.5& 77.4& 4.7& 51.7\\
&MCD \cite{saito2018maximum}& 87.8& \underline{75.7}& \underline{84.2}& \underline{78.1}& 91.6& \underline{95.3}& 88.1& 78.3& 83.4& 64.5& 84.8& 20.9& 77.7\\
&MDD \cite{zhang2019bridging}& 88.3& 62.8& \textbf{85.2}&	69.9& \underline{91.9}&	95.1& \textbf{94.4}&	\textbf{81.2}& \textbf{93.8}&	\underline{89.8}& 84.1& \underline{47.9}& \underline{82.0}\\
\cline{2-15}
&CDAN \cite{long2018conditional}&  \underline{94.0}& 69.2& 78.9& 57.0& 89.8& 94.9& 91.9& \underline{80.3}& 86.8& 84.9& \underline{85.0}& \textbf{48.5}& 80.1\\
&TriDA+CDAN& \textbf{95.7}& \textbf{77.3}& 82.8& 58.2& \textbf{94.2}& \textbf{98.5}& \underline{93.5}& 78.0& \underline{92.2}& \textbf{90.9}& \textbf{89.7}& 45.1& \textbf{83.0}\\
\midrule
\midrule
\multirow{12}{*}{\rotatebox{90}{SFUDA}}
&SFDA \cite{kim2021domain}& 86.9& 81.7& 84.6& 63.9& 93.1& 91.4& 86.6& 71.9& 84.5& 58.2& 74.5& 42.7& 76.7\\
&A$^2$Net \cite{xia2021adaptive}& 94.0& 87.8& 85.6& 66.8& 93.7& 95.1& 85.8& 81.2& 91.6& 88.2& 86.5& 56.0& 84.3\\
&NRC \cite{yang2021exploiting}& \underline{96.8}& \textbf{91.3}& 82.4& 62.4& 96.2& 95.9& 86.1& 80.6& 94.8& 94.1& \underline{90.4}& 59.7& 85.9\\
&HCL \cite{huang2021model}& 93.3& 85.4& 80.7& 68.5& 91.0& 88.1& 86.0& 78.6& 86.6& 88.8& 80.0& \textbf{74.7}& 83.5\\
&DIPE \cite{wang2022exploring}&95.2& 87.6& 78.8& 55.9& 93.9& 95.0& 84.1& 81.7& 92.1& 88.9& 85.4& 58.0& 83.1\\
&AaD \cite{yang2022attracting}& \textbf{97.4}& 90.5& 80.8& 76.2& \textbf{97.3}& 96.1& 89.8& 82.9& 95.5& 93.0& \textbf{92.0}& \underline{64.7}& \underline{88.0}\\
&SFDA-DE \cite{ding2022source}& 95.3& \underline{91.2}& 77.5& 72.1& 95.7& \textbf{97.8}& 85.5&  \underline{86.1}& 95.5& 93.0& 86.3& 61.6& 86.5\\

&DaC \cite{zhang2022divide}& 96.6& 86.8& \textbf{86.4}& \textbf{78.4}& 96.4& 96.2& \textbf{93.6}& 83.8& \textbf{96.8}& \textbf{95.1}& 89.6& 50.0& 87.3\\
 \cline{2-15}
&SHOT \cite{liang2020we}& 94.3& 88.5& 80.1& 57.3& 93.1& 94.9& 80.7& 80.3& 91.5& 89.1& 86.3& 58.2& 82.9\\
&TriDA+SHOT& 96.2& 88.7& 78.3& 53.2& 94.5& 95.7& 80.3& 81.2& 93.0& 89.3& 87.1& 62.2 & 83.3\\
&CoWA \cite{lee2022confidence}& 96.2& 89.7& 83.9& 73.8& 96.4& \underline{97.4}& 89.3& \textbf{86.8}& 94.6& 92.1& 88.7& 53.8& 86.9\\
&TriDA+CoWA& 96.7& 90.6&  \underline{86.2}& \underline{77.2}&  \underline{96.7}& 97.2& \underline{92.0}& 82.6& \underline{96.0}&  \underline{94.7}& 88.6& 61.2& \textbf{88.3}\\
\bottomrule
\end{tabular}}
\label{tab:visda}
\end{table*}





\section{Experiment}

\subsection{Dataset}
To verify the effectiveness of TriDA, we evaluate it in a variety of scenarios, including UDA and SFUDA, over three image recognition datasets as introduced below.

\textit{Office-31} \cite{saenko2010adapting} is a small-scale DA benchmark that contains 4110 images from 31 classes. Those images are divided into three domains: Amazon (A), DSLR (D), and Webcam (W).

\textit{Office-Home} \cite{venkateswara2017deep} is a medium-sized benchmark, which comprises 15,500 images from 65 categories. The dataset consists of four different domains: Art (Ar), ClipArt (Cl), Product (Pr), and Real-World (Rw).

\textit{VisDA-C} \cite{peng2017visda} is a challenging large-scale benchmark for adapting from synthetic to real images. There are 12 categories in each domain. The synthetic domain consists of 152,397 synthetic images generated by 3D rendering and the real domain contains 55,388 real-world images.
 


\subsection{Implementation Details.}
We utilize three backbones, ResNet-50, ResNet-101 \cite{he2016deep}, and ViT-B/16 \cite{dosovitskiy2020image}. Following \cite{liang2020we}, we adopt an FC layer with 256 units for the bottleneck, followed by BatchNorm, and an FC layer for the classifier. For optimization, we use mini-batch Stochastic Gradient Descent (SGD) with a momentum of 0.9 and a weight decay of 1e-3. The learning rate is set at 1e-3 for backbones pre-trained on ImageNet and 1e-2 for the remaining layers, except 1e-3 for VisDA-C. For all experiments, we maintain the value of $\beta=0.1$. We show the class selection settings in \cref{tab:class}. Each experiment is performed three times using different random seeds, with the average accuracy being reported.

\begin{table}
\caption{Ablation study on Office-Home (ResNet50).}

\begin{center}
\begin{tabular*}{\hsize}{@{}@{\extracolsep{\fill}}ccccccccc@{}}
    \toprule
     $\mathcal{L}_p$& $\mathcal{L}_{sem}$& $\mathcal{L}_{feat}$& Cl$\rightarrow$Pr &Cl$\rightarrow$Ar& Cl$\rightarrow$Rw\\
    \midrule
     & & & 78.2 &68.0 &78.0\\
    $\checkmark$ & & & 79.9 &69.4 &80.3\\
     & $\checkmark$& &79.9 &69.2 & 80.4\\
    $\checkmark$ & $\checkmark$& &\underline{81.2} &\underline{69.6}&80.6\\
     & $\checkmark$&$\checkmark$ &{80.7} &69.5&\underline{81.0}\\
    $\checkmark$ & $\checkmark$& $\checkmark$& \textbf{81.3}& \textbf{70.1}& \textbf{82.1}\\
    \bottomrule
    \end{tabular*}
    \vspace{-2mm}
\end{center}
\label{ablation}
\end{table}

\begin{table}
\caption{Analysis of hyperparameter $\alpha$ on Office-Home (ResNet50).}
\begin{center}
\begin{tabular}{ccccc}
    \toprule
     $\alpha$ & 0.5& 1.0& 2.0 & 4.0\\
     \midrule
     Avg. & 73.2& 73.3& 73.5 & 73.3\\
    \bottomrule
    \end{tabular}
    \vspace{-2mm}
\end{center}
\label{tab:alpha}
\end{table}

\subsection{Comparison with State-of-the-arts}
We evaluate \mName in both vanilla UDA and SFUDA scenario, and \mName achieves state-of-the-art performance. Specifically, in \cref{tab:office31}, we summarize classification accuracies on Office-31 dataset. \revise{In UDA, TriDA brings the improvements up to 1.7\% and 1.1\% over CDAN and MDD, respectively. While in SFUDA, TriDA enhances SHOT by 1.8\%. The experiment results on Office-Home dataset are shown in \cref{tab:officehome}. In UDA, TriDA outperforms the baseline, CDAN and MDD , by a margin of 2.1\% and 1.7\%.} Similarly, in SFUDA, TriDA enhances the baseline, SHOT, by 1.8\%. Additionally, we report the results on ViT in \cref{tab:officehome_vit}. The improvement (1.3\% on SHOT and 1.2\% on CoWA) shows the efficacy of our method across different backbones. \revise{The experiment results on VisDA-C dataset are shown in \cref{tab:visda}. UDA/SFUDA methods exhibit imbalanced performance across classes. For instance, CoWA achieves 97.4\% on "knife" but only 53.8\% on "truck," indicating disparities in pseudo-label quality and pre-trained knowledge degeneration. Our method primarily benefits classes where pre-trained knowledge has significantly degraded, such as improving "truck" from 53.8\% to 61.2\%. Conversely, for classes where pre-trained knowledge remains well-preserved and pseudo-labels are of high quality, the improvement is more limited.} In UDA, TriDA achieves top-1 accuracy on multiple tasks, such as plane, bike, and horse. \revise{In SFUDA, TriDA improves performance in 9 out of 12 classes on SHOT and CoWA. In terms of average accuracy, it enhances the baseline, CoWA, by 1.4\%.} These results demonstrate that incorporating pre-training data represents a straightforward yet potent approach to improving UDA performance.

\subsection{Analysis}
\label{analysis}



\noindent\textbf{Ablation study.} We conduct the ablation study of SFUDA on Office-Home to verify the effect of each component in TriDA, including $\mathcal{L}_p$, $\mathcal{L}_{sem}$ and $\mathcal{L}_{feat}$. The results are shown in \cref{ablation}. We observe that only utilizing $\mathcal{L}_{p}$ brings the significant improvement, $1.7\%$ on Cl$\rightarrow$Pr and $1.4\%$ on Cl$\rightarrow$Ar. It demonstrates that without intermediate domain, directly incorporating pre-training data is help to prevent the pre-trained knowledge degeneration and reduce the impact of pre-training, improving the generalization ability of adapted models. Moreover, introducing $\mathcal{L}_{sem}$ and $\mathcal{L}_{feat}$ further improve the performances, confirming that intermediate domain data can alleviate difference between objective function on pre-training data and target domain, and achieve a lower target risk boundary. In addition, removing $\mathcal{L}_{feat}$ degrades the accuracy to $81.2\%$ on Cl$\rightarrow$Pr, which verified their effectiveness. \revise{See supplementary material for more results.} 

\begin{figure}
    \centering
\includegraphics[width=\linewidth]{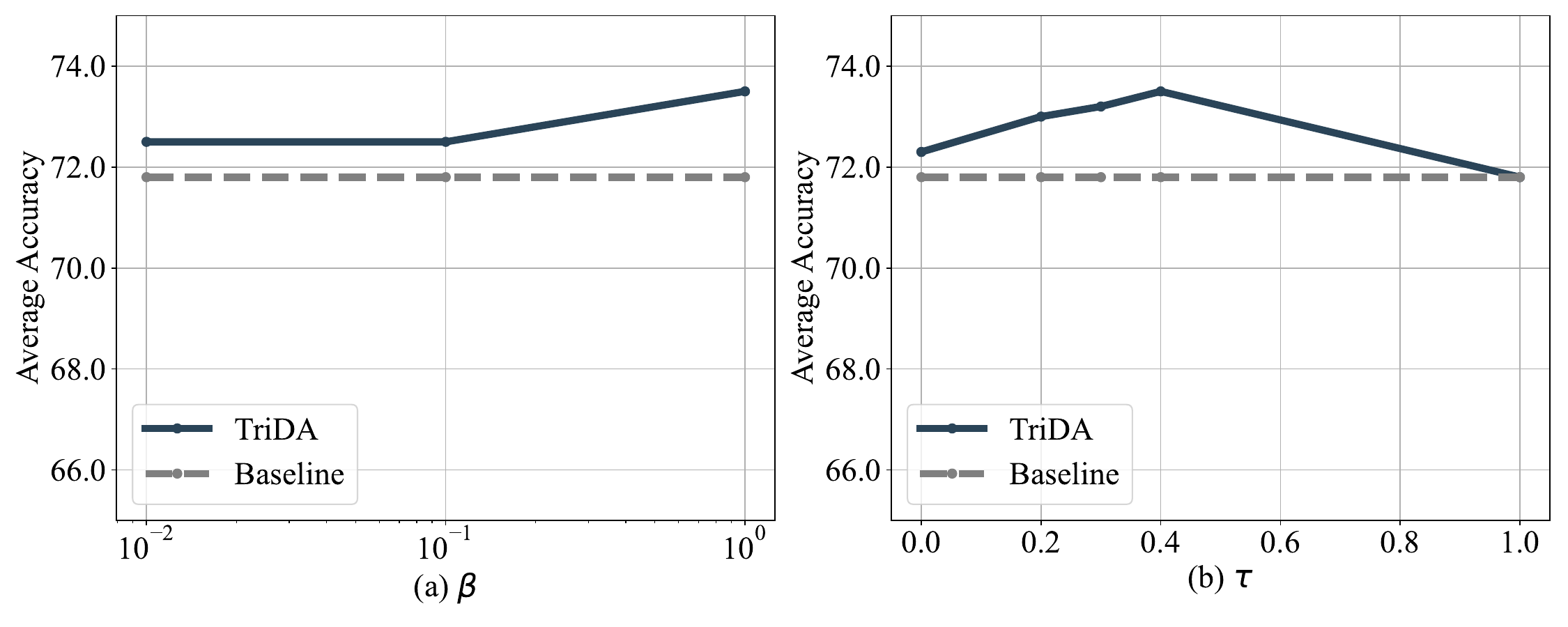}
    \caption{(a)Analysis of hyperparameter loss weight $\beta$ on Office-Home (ResNet50). (b)Analysis of the class similarity threshold $\tau$ on Office-Home (ResNet50).}
    \label{fig:hyperparameter}
\end{figure}

\noindent \textbf{Hyperparameter $\beta$.}
We evaluate the parameter sensitivity of $\beta$ in \cref{betae} (for intermediate domain images). The results with  different $\beta \in \{0.01, 0.1, 1.0\}$ on Office-Home is shown in \cref{fig:hyperparameter}(a). The results demonstrate that our method is effective over a broad range, highlighting its robustness. See supplementary material for detailed results.

\noindent \revise{\textbf{Hyperparameter $\alpha$.}
Since the intermediate domain data is generated based on $\lambda \sim Beta(\alpha, \alpha)$, we assess the sensitivity of our method to the distribution of $\lambda$ by varying $\alpha$ from 0.5 to 4.0. The results, presented in \cref{tab:alpha}, demonstrate that our method remains effective across a broad range of 
$\alpha$, highlighting its robustness. For detailed results, please refer to the supplementary material.}

\begin{table*}
\centering
\caption{Analysis of Class Selection Strategy on Office-Home (ResNet50).}
\small
\resizebox{\linewidth}{!}{
\begin{tabular}{lccccccccccccccc}
\toprule
Method& Ar$\rightarrow$Cl&	Ar$\rightarrow$Pr& Ar$\rightarrow$Rw& Cl$\rightarrow$Ar& Cl$\rightarrow$Pr& Cl$\rightarrow$Rw& Pr$\rightarrow$Ar& Pr$\rightarrow$Cl& Pr$\rightarrow$Rw&	Rw$\rightarrow$Ar& Rw$\rightarrow$Cl& Rw$\rightarrow$Pr& Avg.\\
\midrule
Baseline& {57.1}& {78.1}&  {81.5}&  {68.0}&  {78.2}&  {78.1}&  {67.4}& 54.9&  {82.2}&  {73.3}&  {58.8}&  {84.3}&  {71.8}\\
Random& \textbf{59.3}& 76.7& \underline{82.9}& \underline{69.0}& 78.2& \underline{80.9}& 68.0& \underline{55.0}& 82.5& \textbf{74.9}& 59.4& \underline{84.4}& 72.6&\\
Select All& 57.0& 76.2& 81.6& 68.4& \underline{81.1}& 80.3& 68.2& \textbf{55.3}& \textbf{83.3}& 73.5& {59.7}& 83.5& 72.3\\
UOT\cite{liu2022improved}& \underline{59.0}& \underline{79.8}& 82.7& 68.6& 80.5& \underline{80.9}& \underline{69.5}& \textbf{55.3}& {82.9}& 74.5& \textbf{60.2}& 84.1& \underline{73.2}\\
TriDA &57.3& \textbf{80.2}& \textbf{83.1}& \textbf{70.1} &\textbf{81.3}& \textbf{82.1}& \textbf{70.7}& 54.7& \underline{83.0}& \underline{74.3}&  \underline{59.8}&\textbf{85.7}& \textbf{73.5}\\
\bottomrule
\end{tabular}}
\label{tab:cls}
\end{table*}

\noindent \textbf{How many classes to select? }
We adjust the class similarity threshold, $\tau$, from $0.0$ to $0.4$. The results in \cref{fig:hyperparameter}(b) indicate that selecting merely 63 ($\tau=0.2$) categories can enhance performance. Incorporating all classes achieves suboptimal performance. It suggests that selecting semantically images irrelevant to target domain brings less transferable knowledge and increases target risk. See supplementary material for detailed results. 

\noindent \textbf{Class Selection Strategy.}
We conduct experiments employing various class selection strategies. We select 63 classes for all strategies except the "Select All" strategy, which includes 1000 classes. The results on the Office-Home dataset with SHOT~\cite{liang2020we} as baseline are presented in \cref{tab:cls}. Firstly, compared with selecting a subset (Random, UOT  \cite{liu2022improved}, and TriDA), the performance drops when selecting all classes. Despite this, all strategies achieve higher accuracy than the baseline.  It demonstrates that incorporating more images does not always bring improvement. We hypothesize that this is due to the excessive number of irrelevant images which increases $\delta^2$ in \cref{modif}, resulting in a higher target risk bound. 




\begin{figure}
    \centering
\includegraphics[width=\linewidth]{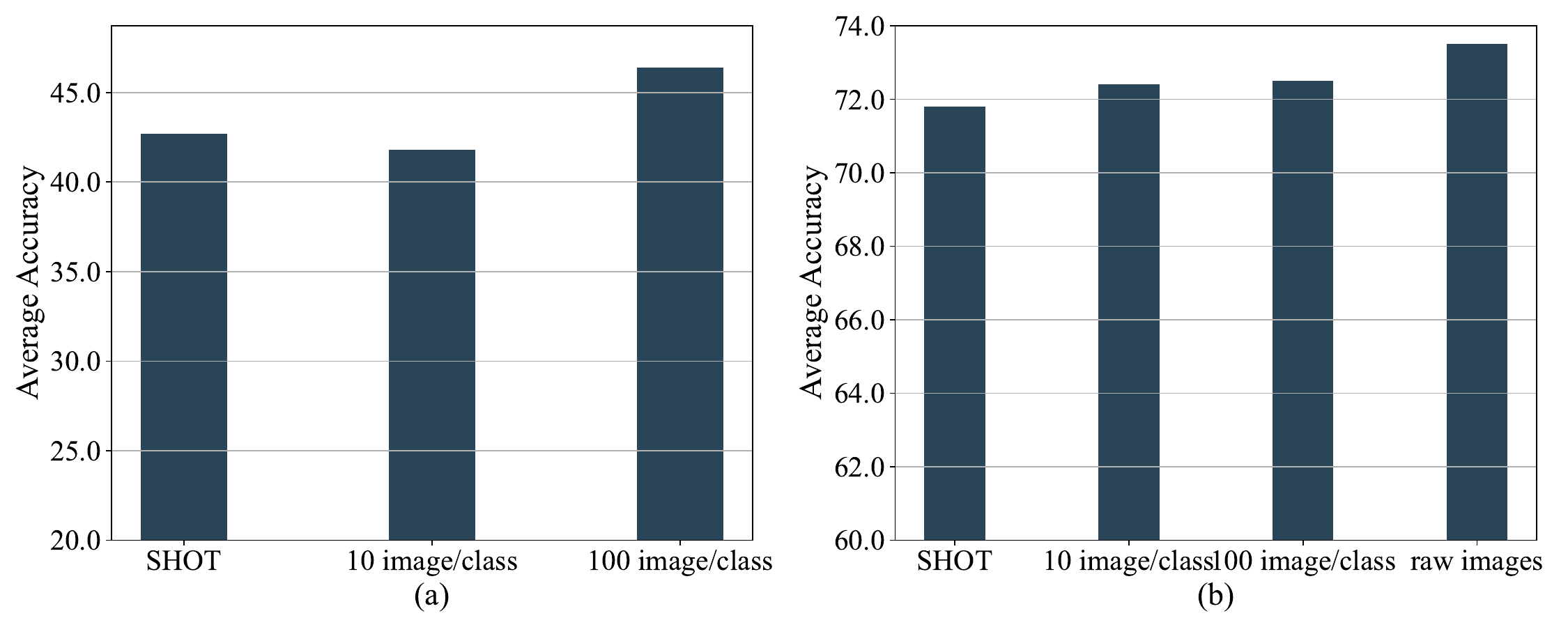}
    \caption{(a)Analysis of the number of synthesized images per class on Office-Home (ResNet50 pre-trained on CLIP). (b) Analysis of the number of synthesized images per class on Office-Home (ResNet50 pre-trained on ImageNet). Note that our method could
bring benefits with only 10 images per class.}
    \label{fig:bar}
\end{figure}

\begin{figure}
    \centering
    \includegraphics[width=0.9\linewidth]{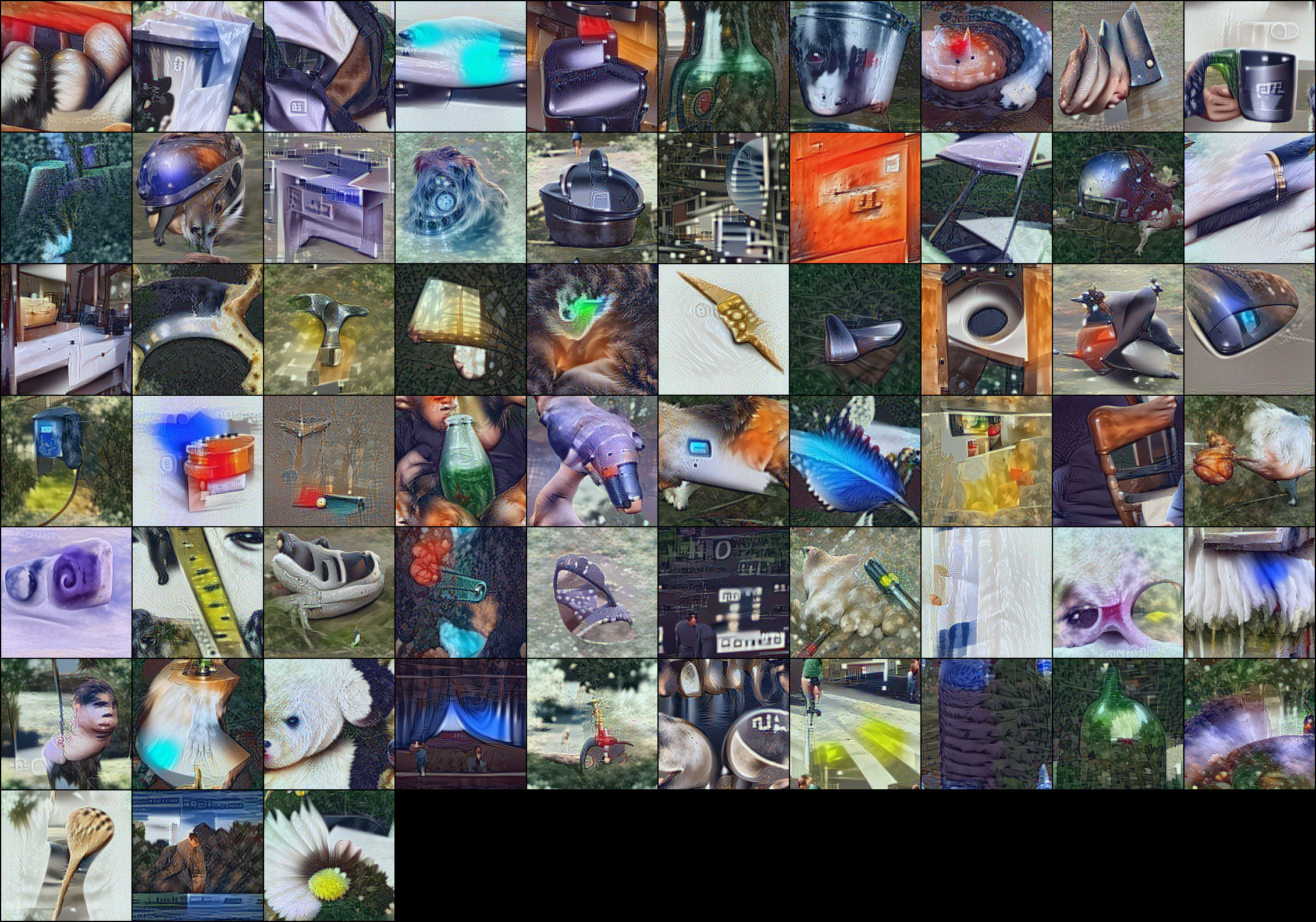}
    \caption{Synthesized images of 63 selected classes with the ImageNet pre-training weight.}
    \label{res}
\end{figure}

\begin{figure}
    \centering
    \includegraphics[width=0.9\linewidth]{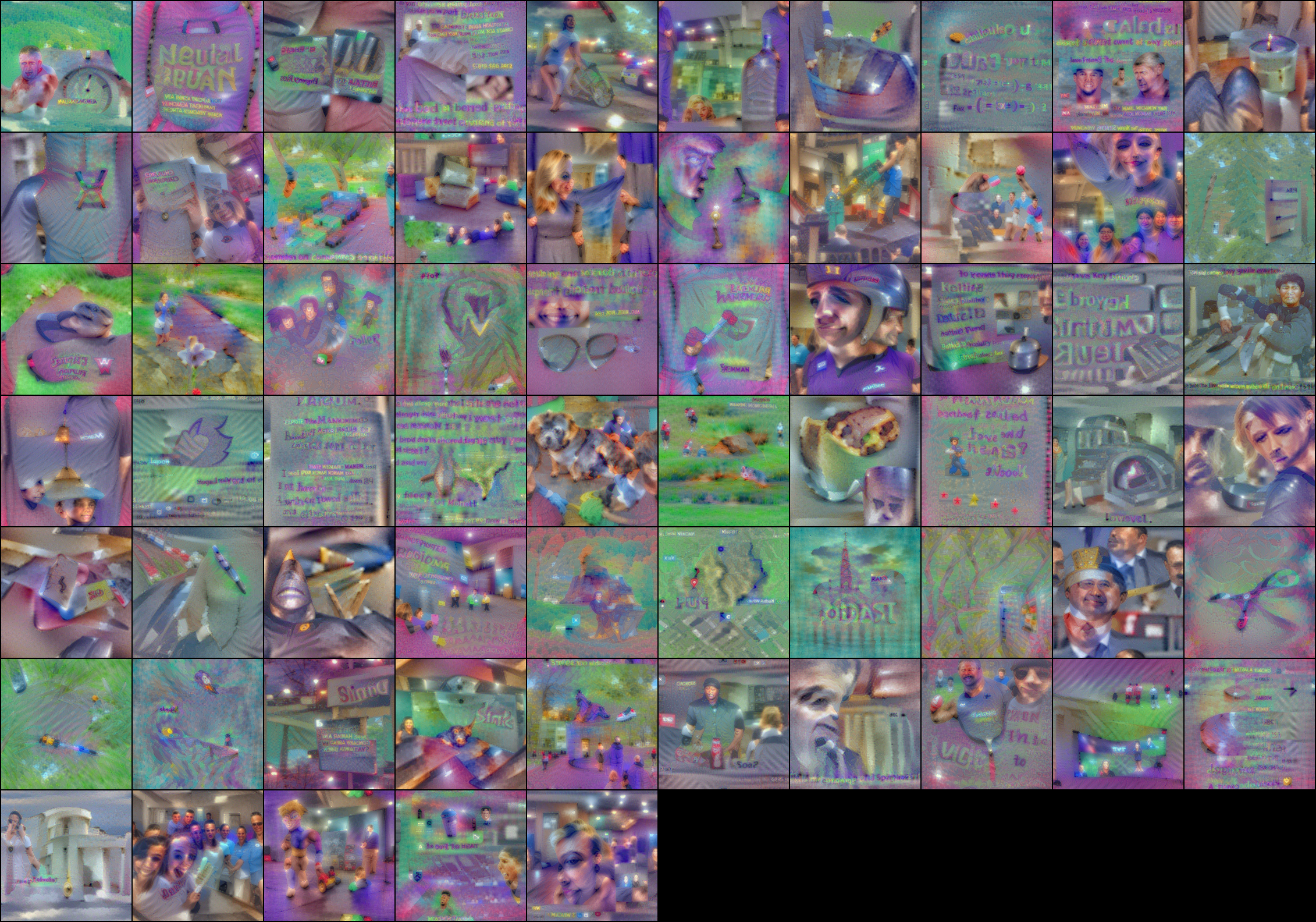}
    \caption{Synthesized images of 65 classes from Office-Home with the CLIP pre-training weight. }
    \label{clip}
\end{figure}

\begin{figure}
    \centering
    \includegraphics[width=\linewidth]{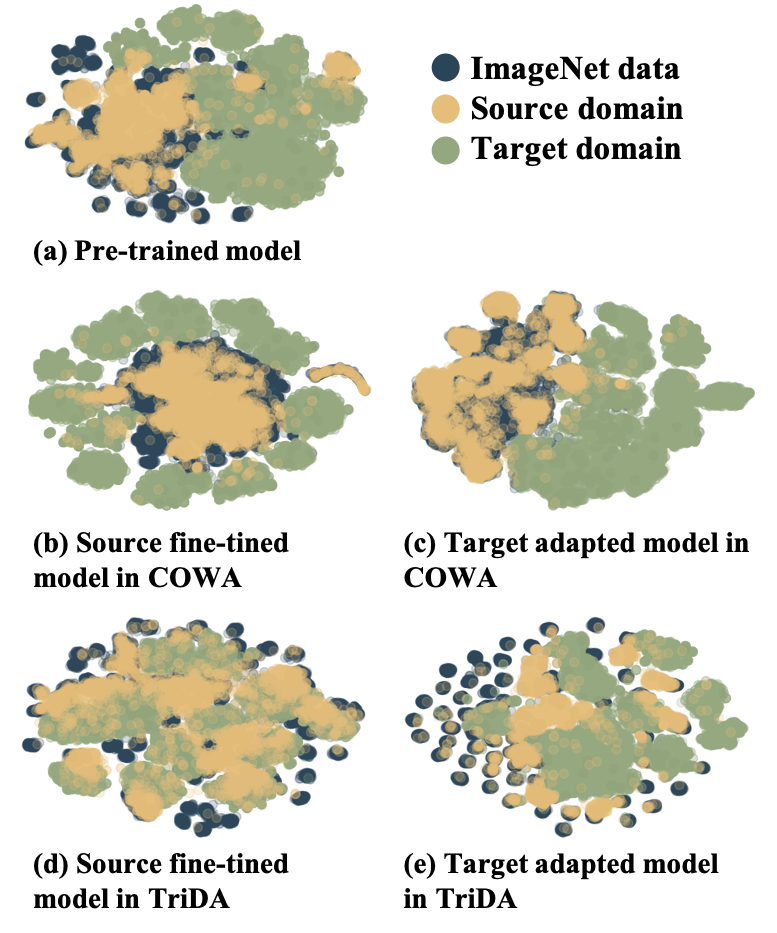}
    \caption{After adaptation, the overlap between the target and pre-training data domains, as well as the overlap between the source and pre-training data domains, increases, implying Pre-training Gravity. Our method preserves the semantic clusters of ImageNet and achieves better alignment between the source and target domains.}
    \label{tsne_all}
\end{figure}

\begin{figure}
    \centering
     \includegraphics[width=\linewidth]{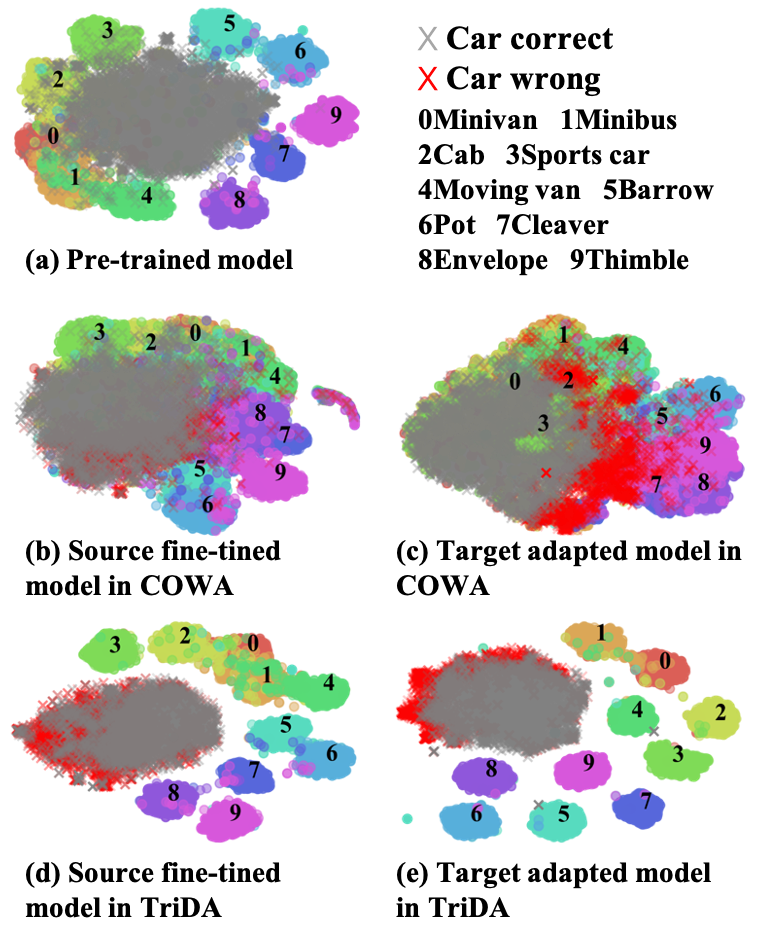}
    \caption{Pre-trained Knowledge Degeneration makes the boundary of ImageNet clusters vanish. Wrong predictions primarily lie in the overlap of unrelated classes in ImageNet. Our method mitigates it by preserving the cluster structures.}
    \label{tsne_car}
\end{figure}


\noindent \textbf{Scalability to Pre-training Methods.} 
In addition to supervised pre-training on ImageNet, we extend our experiments to include the multi-modal pre-training model CLIP \cite{Radford2021LearningTV}.  For adaptation, prompts with the target dataset’s classes are embedded by the fixed text encoder to synthesize the classifier. We reproduce the comparison method, and the result on Office-Home is shown in \cref{fig:bar}(a). TriDA surpasses the baseline by using only 6500 synthesis images in total, showing its effectiveness across different pre-training methods.

\begin{table}[h]
\small
\caption{Classification accuracy (\%) on medical imaging (ResNet50).}
\begin{center}

\resizebox{\linewidth}{!}{\begin{tabular}{lcccccccccccc}
    \toprule
     Method &C1$\rightarrow$C2  &C1$\rightarrow$Cy  &C2$\rightarrow$C1 &C2$\rightarrow$Cy  &Cy$\rightarrow$C1 &Cy$\rightarrow$C2 & Avg. \\
     \midrule
     ResNet50 &  63.6& 43.6 & 87.4 &61.9  &46.8 &48.1 & 58.6\\
     SHOT\cite{liang2020we}&  \textbf{98.5}& 43.9&  96.9& \textbf{90.9}& 51.8 &70.0 &75.3\\
     TriDA+SHOT\cite{liang2020we}& \textbf{98.5}& \textbf{60.0} & \textbf{97.7}& 90.2&  \textbf{52.9}& \textbf{74.8}& \textbf{79.0} \\
    \bottomrule
    \end{tabular}}
\end{center}
\label{tab:colon}
\end{table}



\noindent \textbf{How many images to synthesize? } With synthesized images, TriDA demonstrates significant improvements on Office-Home, as depicted in \cref{fig:bar}(b). Notably, even with as few as 10 synthesized images per class, TriDA shows benefits on ImageNet, with performance continuing to increase with more images. The results show the effectiveness of our method when the raw images are unavailable. In addition, the analysis on CLIP is shown in \cref{fig:bar}(a). With 10 images per class, we observe a lower average performance. Increasing the number of images to 100, we can observe that our method significantly outperforms the baseline. The results show the effectiveness of our method across different pre-training weights. See supplementary material for detailed results.

\noindent \textbf{Visualization of synthesized images.}
We visualize the synthesized images from the ImageNet pre-trained weight and CLIP in \cref{res} and \cref{clip}, respectively. We observe that the synthesized images have lower fidelity than ImageNet. We conjecture this is because the CLIP dataset contains images with complex scenarios rather than a single object. Thus, CLIP needs more synthesized images to improve the adaptation performances.

\noindent \textbf{Visualization of features.} We first depict t-SNE \cite{van2008visualizing} in \cref{tsne_all}. Compared with the pre-trained model, the enlarged overlap after adaptation among the three domains implies \textit{Pre-training Gravity}. Meanwhile, \textit{Pre-trained Knowledge Degeneration} makes the boundary of ImageNet clusters vanish. \revise{In \cref{tsne_all} (c) and (e), our method effectively prevents pre-trained knowledge degeneration, resulting in clear cluster boundaries for ImageNet features, and achieves better alignment between the source and target domain. Since ImageNet and the source/target datasets share the same or similar classes, the source/target data are attracted by ImageNet data due to pre-training gravity. This leads to overlap in the feature space, meaning that semantically similar features, regardless of the domain, lie close to each other.} \cref{tsne_car} illustrates a more specific examples. \texttt{"Car"} (in grey and red) are close to \texttt{"Minivan"} and \texttt{"Minibus"} from ImageNet due to semantic similarity. In CoWA, we observe that wrong target predictions (red points) primarily lie on the overlap part with unrelated classes of ImageNet (red cross near \texttt{"Cleaver"}). This is because the target features tend to align with impaired ImageNet's cluster structures due to the pre-training Gravity. In contrast, our method mitigates it by preserving the cluster structure of ImageNet.

\noindent \textbf{\revise{Application in Medical Imaging.}}
\revise{UDA datasets and pre-training datasets often share significant label space overlap. To further investigate the scalability of TriDA, we evaluate its performance on two colon histopathological image datasets—Camelyon17\cite{bandi2018detection} and Chaoyang\cite{zhu2021hard}—in which images are labeled as either adenocarcinoma or normal. The datasets comprise three domains: two hospitals from Camelyon17, denoted as C1 and C2, and Chaoyang, denoted as Cy. Although the pre-training dataset does not contain classes that directly correspond to those in the target datasets, we reuse all classes from the pre-training dataset. The results in \cref{tab:colon} indicate that TriDA achieves a significant improvement, with average accuracy increasing from 75.3\% to 79.0\%. We attribute this enhancement to the preservation of low-level knowledge, such as color and texture, which is beneficial for domain adaptation. This finding demonstrates the generalizability of TriDA under conditions where semantic overlap is minimal, highlighting its potential for practical applications.}

\section{Conclusion} 
This paper provides a comprehensive study of the pre-training dataset in UDA through both empirical and theoretical analysis. We reveal the target error stemming from pre-training data which is not considered by the two-domain paradigm. Thus, we propose a novel framework TriDA, incorporating the pre-training data into UDA and redefining it as a three-domain problem. We believe our work sheds light on the role of pre-training in UDA, contributing to a deeper understanding of this field and paving the way for more robust and effective domain adaptation.

\textbf{Viability.} Firstly, our method takes into consideration the availability of pre-training data and provides a solution using synthesized images. Secondly, although the pre-training dataset is always large-scale, our method offers benefits with minimal pre-training or synthesized images. As demonstrated in \cref{analysis}, utilizing 75k pre-training images (1/3 the size of VisDA-C) improves accuracy, and incorporating 10 synthesized images per class (650 images total) enhances accuracy on the office-home. Thirdly, our method only adds a lightweight classifier during training, which is removed during inference. Therefore, no extra memory or computation time is required for inference.

\textbf{Scalability.} Our method is plug-and-play and seamlessly complements existing UDA and SFUDA methods with improvements across different backbones without additional inference costs. Additionally, our method can be integrated into various pre-training methods, including supervised learning and natural language supervision (such as CLIP).

\textbf{Limitations.} A limitation of TriDA is its reliance on the semantic similarity between the pre-train dataset and the target domain. When the target domain significantly deviates from the pre-train one, the performance needs further investigation.

\IEEEpubidadjcol

\bibliographystyle{IEEEtranN}
\bibliography{ref}

\begin{IEEEbiography}[{\includegraphics[width=1in,height=1.25in,clip,keepaspectratio]{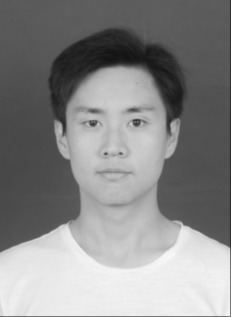}}]{Yinsong Xu} received the bachelor’s degree in communication engineering from the Beijing University of Posts and Telecommunications. He is working toward the PhD degree at the Beijing University of Posts and Telecommunications. Currently, he is also affiliated with the National Institute of Health Data Science, Peking University.
\end{IEEEbiography}

\begin{IEEEbiography}[{\includegraphics[width=1in,height=1.25in,clip,keepaspectratio]{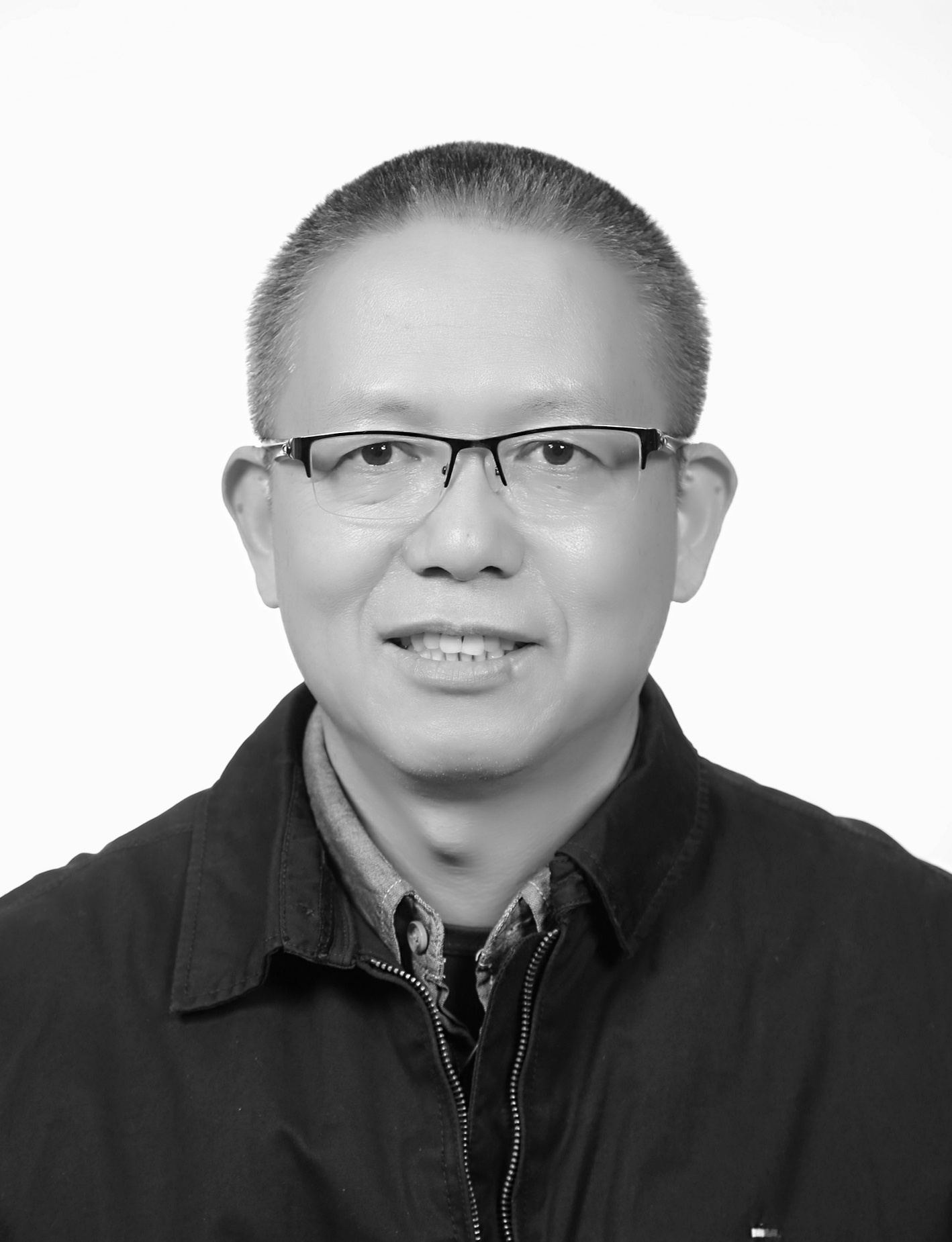}}]
{Aidong Men} is a professor at the School of Artificial Intelligence, Beijing University of Posts and Telecommunications. His research interests include multimedia communication, digital TV, and images and speech signal processing and transmission. Men is a Fellow of the Chinese Institute of Electronics and China Institute of Communications. He is also an Invited Fellow of the Science and Technology Committee of State Administration of Radio, Film, and Television.
\end{IEEEbiography}

\begin{IEEEbiography}
[{\includegraphics[width=1in,height=1.25in,clip,keepaspectratio]{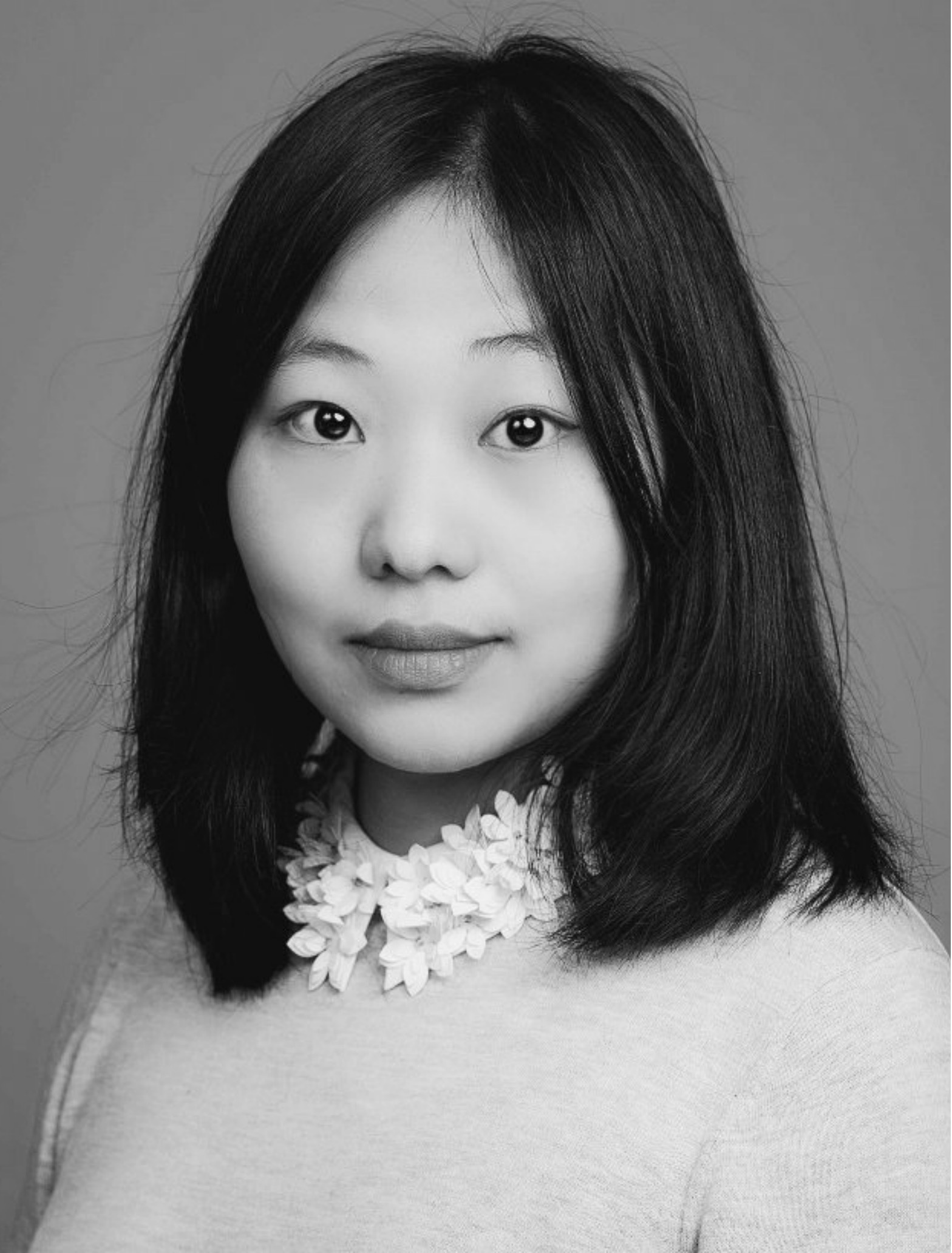}}]
{Yang Liu} received PhD and MPhil in Advanced Computer Science from University of Cambridge, and B.Eng. in Telecommunication Engineering from Beijing University of Posts and Telecommunications (BUPT). She is now a Tenure-track Assistant Professor (Ph.D. Supervisor) in Wangxuan Institute of Computer Technology, Peking University.
\end{IEEEbiography}

\begin{IEEEbiography}
[{\includegraphics[width=1in,height=1.25in,clip,keepaspectratio]{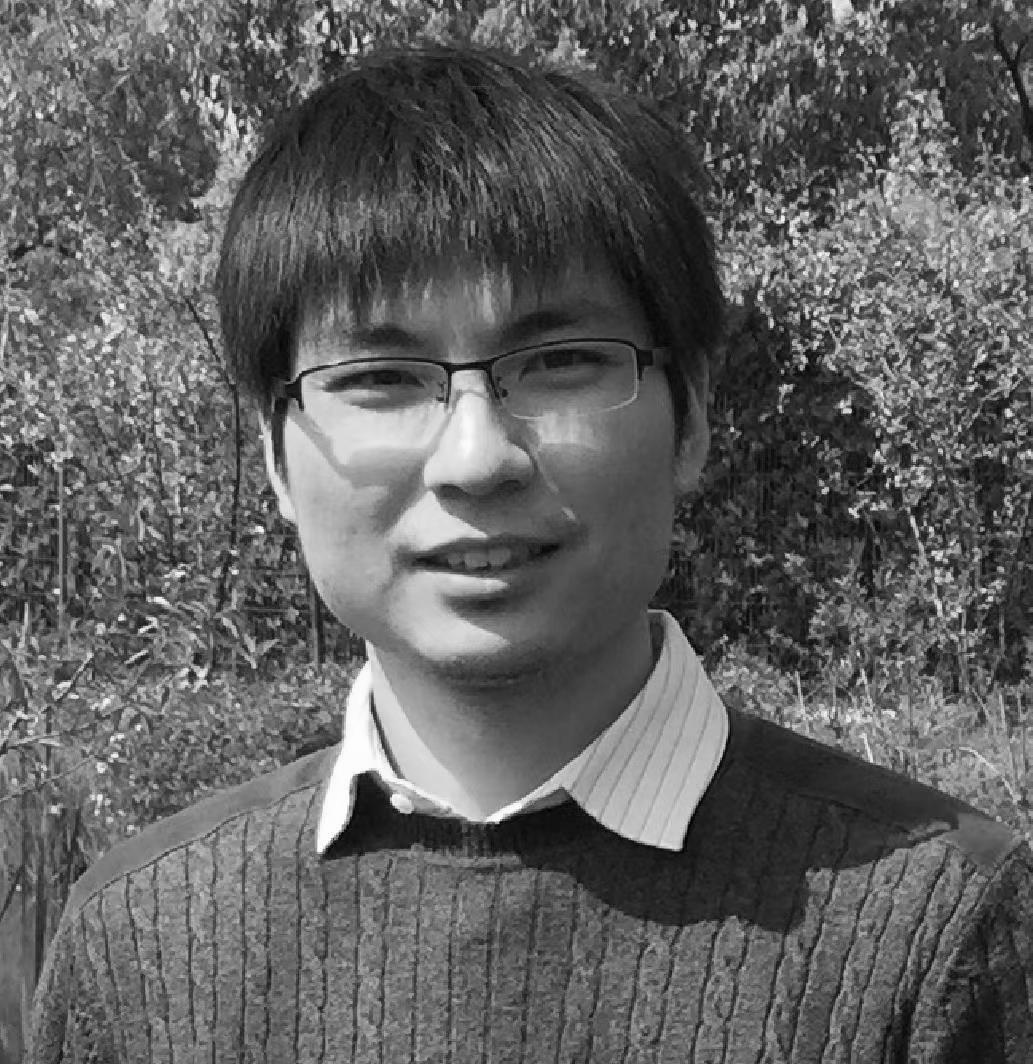}}]
{Xiahai Zhuang} is a professor at the School of Data Science, Fudan University. He graduated from Department of Computer Science, Tianjin University, received Master degree from Shanghai Jiao Tong University and Doctorate degree from University College London. His research interests include interpretable AI, medical image analysis and computer vision. His work won the Elsevier-MedIA 1st Prize and Medical Image Analysis MICCAI Best Paper Award 2023.
\end{IEEEbiography}

\begin{IEEEbiography}
[{\includegraphics[width=1in,height=1.25in,clip,keepaspectratio]{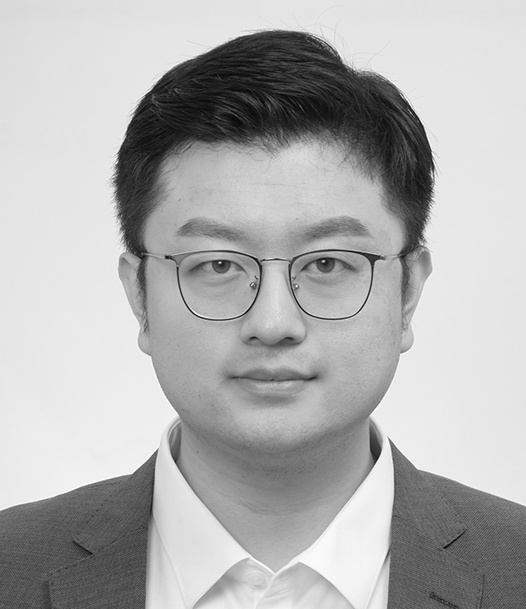}}]
{Qingchao Chen} received the B.Sc. degree in telecommunication engineering from the Beijing University of Post and Telecommunication and the Ph.D. degree from the University College London. He was a Postdoctoral Researcher with University of Oxford, U.K (2018-2021). He is currently a assistant professor the National Institute of Health Data Science, Peking University. His current researches focus on computer vision and machine learning, radio-frequency signal processing and system design, and biomedical multimodality data analysis.
\end{IEEEbiography}

\end{document}